\documentclass[11pt]{article}
\usepackage{amssymb}
\newcommand{\m}[1]{\texttt{\detokenize{#1}}}
\usepackage[utf8]{inputenc}
\usepackage[T1]{fontenc}
\usepackage{graphicx}
\usepackage{booktabs}
\usepackage{multirow}
\usepackage[]{acl}
\newcommand{\cmark}{\ensuremath{\checkmark}}  
\newcommand{\xmark}{\ensuremath{\times}}
\usepackage{times}
\usepackage{latexsym}
\usepackage{amsmath}
\usepackage{color,colortbl}
\usepackage{comment}
\usepackage[T1]{fontenc}

\usepackage[utf8]{inputenc}

\usepackage{microtype}

\usepackage{inconsolata}

\usepackage{graphicx}
\usepackage{xspace}
\definecolor{Gray}{gray}{0.85}

\newcommand*{\afrimteb}{AfriMTEB}

\newcommand*{\mteb}{MTEB}

\newcommand*{\afrimteblite}{AfriMTEB-Lite}

\newcommand*{\afriefive}{\textit{AfriE5-Large-Instruct}}
\newcommand*{\mefive}{\textit{mE5-Large-Instruct}}
\title{Towards Better Text Embeddings for African Languages: Benchmarking and Adaptation via Cross-Lingual Contrastive Distillation}
\title{AfriMTEB and AfriE5: Benchmarking and Adapting Text Embedding Models for African Languages }

\author{
  Kosei Uemura$^{1}$\thanks{Work done during internship at Mila} \quad
  Miaoran Zhang$^{3}$ \quad
  \textbf{David Ifeoluwa Adelani}$^{2,4}$ \\
  \\
  $^{1}$University of Toronto \quad
  $^{2}$Mila - Quebec AI Institute, McGill University \\
  $^{3}$Saarland University, Saarland Informatic Campus \quad
 $^{4}$Canada CIFAR AI Chair \\
  \texttt{k.uemura@mail.utoronto.ca} \\
  \texttt{mzhang@lsv.uni-saarland.de} \\
  \texttt{david.adelani@mila.quebec}
}

\begin{document}
\maketitle


\begin{abstract}
Text embeddings are essential building components of many NLP tasks such as retrieval and clustering. Despite the recent release of the Massive Multilingual Text Embedding Benchmark (MMTEB), African languages remain significantly under-represented.
In this work, we introduce AfriMTEB, a regional extension of MMTEB covering 59 languages, 14 tasks, and 38 datasets. Unlike MMTEB, where many tasks include few or no African languages, AfriMTEB substantially expands coverage, with tasks spanning between 2 and 56 African languages. To address uneven task–language coverage and enable fair evaluation, we further introduce AfriMTEB-Lite, a balanced subset that uniformly covers nine African languages across all tasks.
Complementing the benchmark, we present AfriE5, an adaptation of the instruction-tuned mE5 model to African languages through cross-lingual contrastive learning. Our experimental results show that AfriE5 achieves the strongest overall macro-average among open-weight embedding models on AfriMTEB, with statistically significant gains on several task families, and is competitive with proprietary models such as Gemini Embedding-001.\footnote{The code is publicly available at \url{https://github.com/LLMforLRL/FlagEmbedding-AfriE5}.}

\end{abstract}

\section{Introduction}

\label{sec:intro}

Text embeddings are core building blocks for NLP systems in information retrieval, clustering, semantic similarity, and classification \citep{gao2022simcsesimplecontrastivelearning,feng2022languageagnosticbertsentenceembedding}. However, evaluations on diverse tasks are often limited to a few high resource languages such as English~\citep{muennighoff2023mtebmassivetextembedding} or Chinese~\citep{xiao2024cpackpackedresourcesgeneral}. Many under-represented languages are excluded due to a lack of datasets or non-discoverability of community-created benchmarks~\citep{ojo2025afrobenchgoodlargelanguage}.  

\begin{figure}[t]
    \centering
    \includegraphics[width=\linewidth]{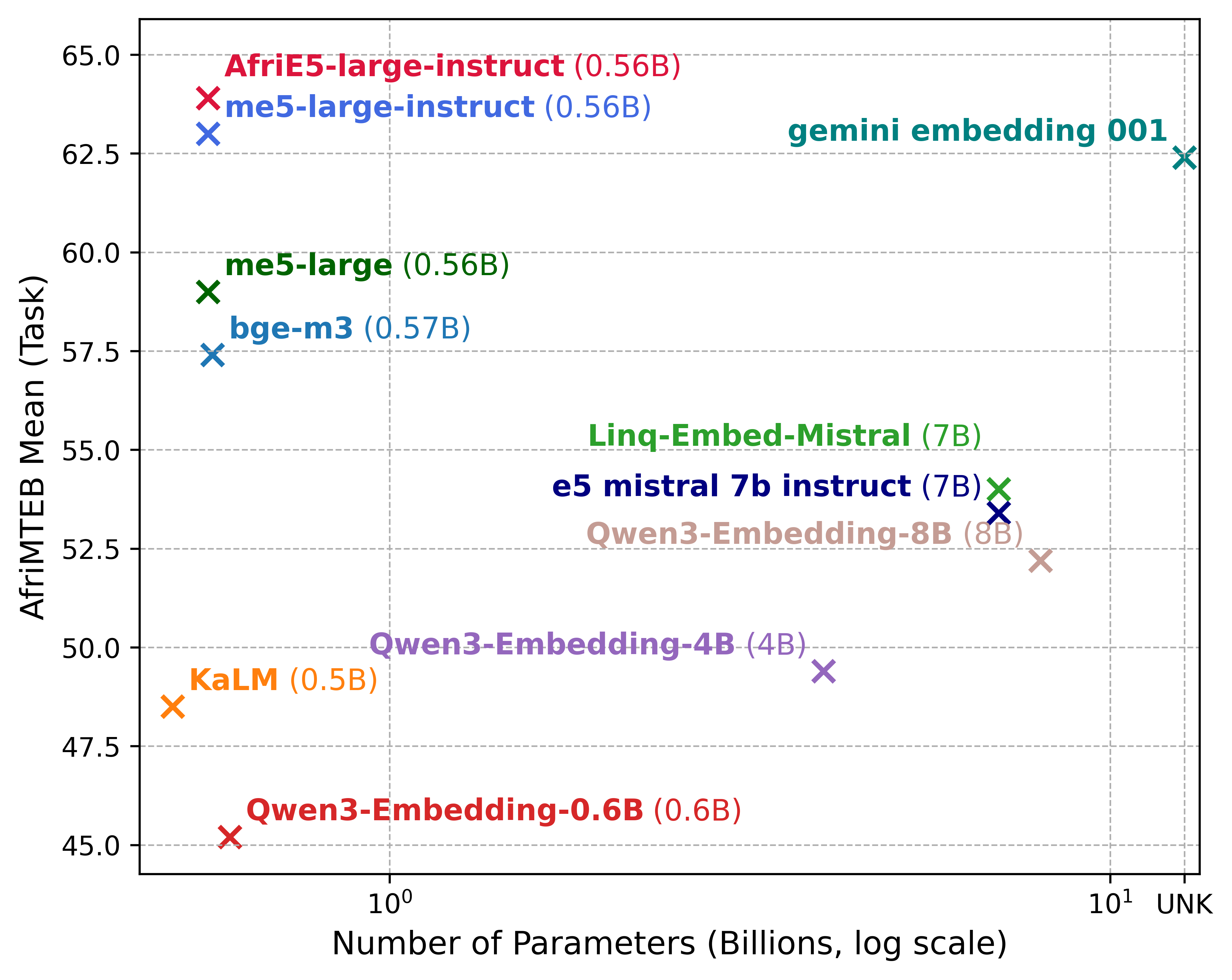}
    \caption{\textbf{Model size vs. mean performance on AfriMTEB.} Parameter counts (billions, log scale) are shown on the $x$-axis, and mean scores across AfriMTEB-Full tasks on the $y$-axis. \textit{AfriE5-large-instruct} achieves the best overall performance (64.6) despite having far fewer parameters than many models.}

    \label{fig:afrimteb-full-score}
\end{figure}

\begin{figure*}[h]
    \centering
    \includegraphics[width=\linewidth]{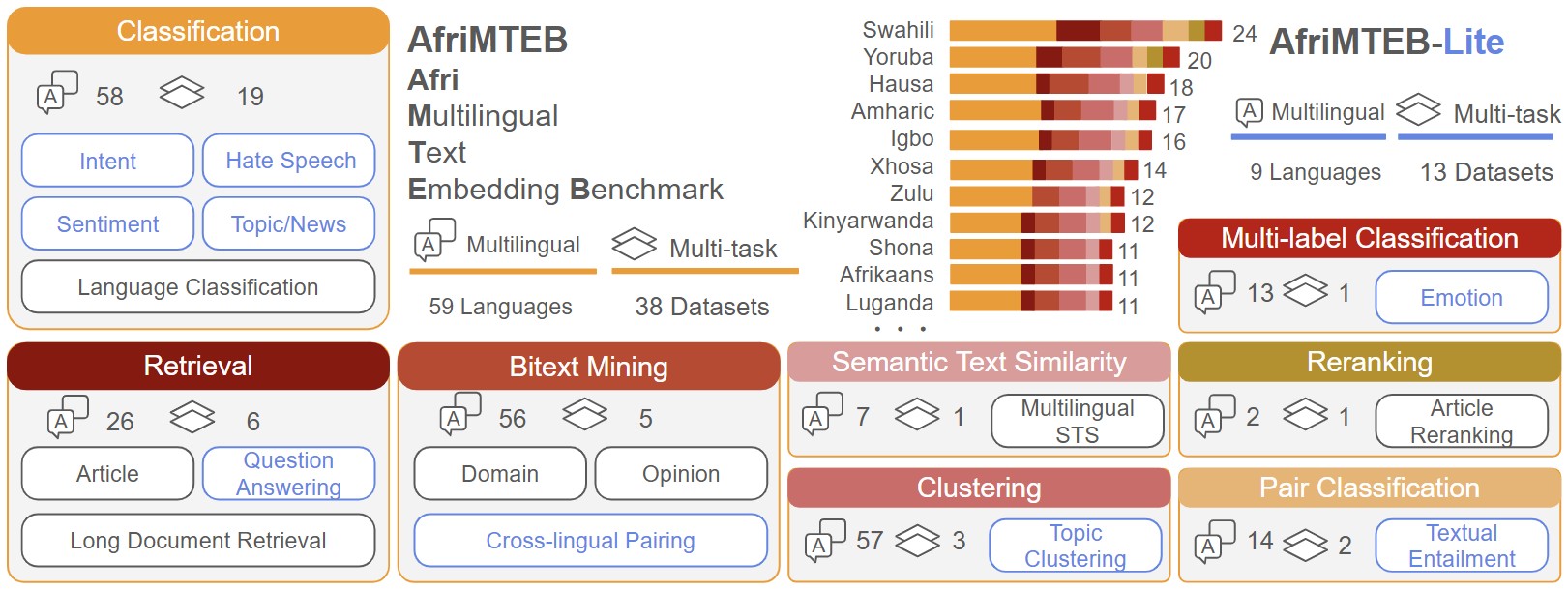}
    \caption{Overview of AfriMTEB and AfriMTEB-Lite. The Full suite spans 59 languages and 38 datasets across 7 families; the Lite suite provides uniform coverage for 9 languages and 13 datasets.}
    \label{fig:afrimteb-overview}
\end{figure*}

While recent text embedding benchmarks have improved language coverage in recent years, such as MMTEB~\citep{enevoldsen2025mmteb}, African languages remain under-represented, where many of the tasks covered are either based on massive evaluation of translation datasets~\citep{nllbteam2022languageleftbehindscaling,federmann-etal-2022-ntrex}, and the tasks derived or repurposed from translation benchmarks such as Belebele~\citep{bandarkar-etal-2024-belebele} and SIB-200~\citep{adelani2024sib200simpleinclusivebig}. As a result, the quality of text embeddings for languages in the African region remains unknown, as the region has few standardized tools for comparing models across tasks and languages~\citep{alabi2025charting}.


In this paper, we introduce \textbf{\afrimteb}, a standardized benchmark designed to evaluate text embeddings for African languages across diverse tasks and application settings. AfriMTEB addresses the lack of task-diverse and systematically comparable evaluations for this region by consolidating and extending existing resources into a unified benchmark. As shown in \autoref{fig:afrimteb-overview}, the \emph{Full} suite (i.e., AfriMTEB-Full) covers 59 languages and 38 datasets spanning bitext mining, classification (single-label, pair, and multi-label), semantic text similarity, retrieval, clustering, and reranking.

To enable controlled, fair, and compute-efficient evaluation, we additionally introduce \textbf{\afrimteblite}, a compact suite over nine geographically and typologically diverse African languages (Amharic, Oromo, Igbo, Yoruba, Hausa, Swahili, Kinyarwanda, Xhosa, and Zulu). AfriMTEB-Lite is constructed by selecting, within each task family, datasets with maximal language overlap, ensuring that almost every task includes all nine languages. This design mitigates the uneven task–language coverage present in large multilingual benchmarks and supports reliable comparison, ablation, and statistical analysis. Crucially, AfriMTEB-Lite enables fine-grained analysis across languages and tasks, allowing us to examine where models succeed or fail by language, resource level, and task family—analyses that are difficult to conduct reliably under the uneven coverage of existing multilingual benchmarks.

In addition to benchmarking, we develop \textbf{\afriefive}, adapting strong embedding models to African languages. 
Starting from the instruction-tuned \textit{mE5-Large-Instruct}~\citep{wang2024textembeddingsweaklysupervisedcontrastive,wang2024multilinguale5textembeddings}, we leverage cross-lingual contrastive learning with knowledge distillation . We construct a contrastive learning dataset by translating MNLI and SNLI datasets~\citep{williams2018broadcoveragechallengecorpussentence,bowman2015largeannotatedcorpuslearning} in African languages with NLLB-200 (3.3B) \citep{nllbteam2022languageleftbehindscaling}, followed by automatic filtering by SSA-COMET~\citep{li2025ssacometllmsoutperformlearned}, an African COMET-based quality estimation (QE) metric. Each example is expanded into multiple sources or target directions to encourage cross-lingual alignment. Additionally, we use BGE Re-ranker v2 m3~\citep{chen2024bgem3embeddingmultilingualmultifunctionality} to extract teacher scores for knowledge distillation. 

We evaluate popular text embedding models, such as BGE-M3~\citep{chen2024bgem3embeddingmultilingualmultifunctionality}, mE5~\citep{wang2024multilinguale5textembeddings}, Qwen embeddings~\citep{zhang2025qwen3embeddingadvancingtext}, Gemini embedding-001~\citep{lee2025geminiembeddinggeneralizableembeddings}, and Embedding Gemma~\citep{vera2025embeddinggemmapowerfullightweighttext}. On \afrimteblite{}, our adapted model \afriefive{}, trained only on nine African languages, achieves an average score of 63.7, surpassing mE5-large-instruct (62.0) and Gemini embedding-001 (63.1). Remarkably, despite being tuned with data and languages aligned to the \afrimteblite, our model leveraged cross-lingual transfer to generalize to the full \afrimteb{} benchmark covering 59 languages and 38 datasets, where it also delivers the best performance with an average score of 62.4, ahead of mE5-large-instruct (61.3) and Gemini embedding-001 (60.6) as shown in \autoref{fig:afrimteb-full-score}. These results highlight that targeted cross-lingual adaptation on a carefully selected subset of languages can transfer effectively to a much larger set, yielding consistent improvements across task families while preserving the backbone’s general utility. 

\section{AfriMTEB Benchmark}

\begin{table*}[t]
\centering
\footnotesize
\setlength{\tabcolsep}{6pt}
\resizebox{\textwidth}{!}{
\begin{tabular}{lllrr}
\toprule
\multirow{2}{*}{\textbf{Task}} 
& \multirow{2}{*}{\textbf{Task Family}} 
& \multirow{2}{*}{\textbf{Datasets}} 
& \multicolumn{2}{c}{\textbf{Number of Languages}} \\
\cmidrule(lr){4-5}
&  &  &  \textbf{MMTEB} & \textbf{AfriMTEB} \\
\midrule
Bitext Mining & Btxt &
\textit{Flores, NTREX, BibleNLP, NollySenti, Tatoeba} & 59 & 59 \\
\addlinespace
\rowcolor{Gray}
NLI & Pr~Clf & \textit{XNLI, \textbf{AfriXNLI}} & 1 & 15 \\
\addlinespace
\rowcolor{Gray}
General Topic class. & Clf & \textit{SIB200Classification, \textbf{SIB200\_14Classes} } & 0 & 56 \\
\rowcolor{Gray}
News Topic class. & Clf & \textit{MasakhaNEWS, TswanaNews, SiswatiNews, SwahiliNews} & 16 & 17 \\
\rowcolor{Gray}
&  & \textit{IsiZuluNews, \textbf{KinNews}} & & \\
Sentiment & Clf & \textit{NaijaSenti, AfriSenti, MultilingualSentiment} & 11 & 12 \\ \rowcolor{Gray}
Hate speech & Clf & \textit{\textbf{AfriHate}} & 0 & 14 \\ 
LID & Clf & \textit{LanguageClassification, SouthAfricanLangClassification} & 0 & 18 \\
& Clf & \textit{AfriSentiLangClassification} &  &  \\ 
\rowcolor{Gray}
Intent & Clf & \textit{MassiveIntent, \textbf{InjongoIntent} }& 3 & 16 \\
Scenario & Clf & \textit{MassiveScenario} & 0 & 3 \\
\addlinespace
\rowcolor{Gray}
Emotion & Multi. Clf & \textit{\textbf{EmotionAnalysisPlus}} & 0 &  14 \\
\addlinespace
Semantic Relatedness & STS & \textit{SemRel24STS} & 7 & 7 \\
\addlinespace
Retrieval & Rtrvl & \textit{Belebele, MIRACL, MIRACLRetrievalHardNegatives,} & 31 & 31 \\
& & \textit{MrTidy, XQuAD, XM3600T2I} & & \\
\addlinespace
Clustering & Clust & \textit{SIB200ClusteringFast, MasakhaNEWSClusteringP2P} & 58 &  58 \\
& & \textit{MasakhaNEWSClusteringS2S} \\
\addlinespace
Reranking & Rrnk & \textit{MIRACLReranking} & 0 & 2  \\
\bottomrule
\end{tabular}
}
\caption{\textbf{Tasks, task families and datasets included in AfriMTEB-Full}. The task families are Bitext Mining (Btxt), Pair Classification (Pr~Clf), Classification (Clf), Semantic Text Similarity (STS), Multi-label Classification (Multi. Clf),  Retrieval (Rtrvl), Clustering (Clust) and Reranking (Rrnk).  We introduce additional the six datasets highlighted in bold. Each dataset is introduced in Appendix \ref{subsec:afrimteb-dataset-descriptions}.}
\label{tab:afrimteb-families}
\vspace{-2mm}
\end{table*}


\afrimteb{} is a region-specific benchmark designed to evaluate text embeddings for African languages across a broad range of tasks. While existing multilingual benchmarks such as MMTEB provide valuable global coverage, their African-language representation is sparse and highly uneven. Several task families central to real-world embedding use—such as hate speech detection, intent classification, and multi-label emotion analysis—include \emph{zero} African languages in MMTEB, while core classification benchmarks often cover only one to three African languages (e.g., XNLI). In contrast, AfriMTEB consolidates and extends existing resources into a standardized evaluation suite covering 59 languages, 14 tasks, and 38 datasets, including six newly introduced datasets that fill critical gaps in task and language coverage.

\afrimteb{} follows the \mteb{}~\citep{muennighoff2023mtebmassivetextembedding} taxonomy and groups tasks into eight families. \autoref{tab:afrimteb-families} lists the tasks, task families and the datasets included in the \emph{Full} suite. Each dataset belongs to one of the eight task families: (1) Bibtext Mining (Btxt), (2) Pair Classification (Pr Clf), (3) Semantic Text Similarity (STS), (4) Clustering (Clust), (5) Classification (Clf), (6) Multi-label Classification (Multi. Clf), (7) Retrieval (Rtrvl), (8) Reranking (Rrnk). The first four task families perform a task based on a pair of sentences, such as identifying translation pairs, recognizing textual entailment, or measuring semantic relatedness. The next two task families (5–6) focus on classifying a sentence or document into one or more categories. Finally, the last two task families (7–8) are information retrieval tasks, either retrieving relevant items based on a query or re-ranking retrieved results. We provide a full description in Appendix \ref{subsec:afrimteb-families}.


\subsection{Existing MMTEB Datasets}

\afrimteb{} builds on the MMTEB by selecting tasks that include African languages. From the original benchmark, we inherit datasets covering bitext mining (e.g., Flores, NTREX, Tatoeba), pair classification (XNLI), topic and sentiment classification (SIB-200, AfriSenti, MasakhaNEWS), semantic textual similarity (SemRel24STS), retrieval (MIRACL, XQuAD, XM3600), clustering (SIB-200, MasakhaNEWS), and reranking (MIRACL).  


However, language coverage in these datasets is uneven: some tasks include only a few African languages, while others span broader multilingual settings. To ensure fairness, in the evaluation, we compute macro-averages over languages within each task, then average across tasks and families to obtain the overall \afrimteb{} score. This prevents any single task or language from dominating the benchmark.

\subsection{New Datasets in AfriMTEB}



In AfriMTEB, we introduce six new datasets to broaden task coverage, increase difficulty, and improve language coverage. Details of these additional datasets are provided below.


\paragraph{AfriXNLI}
An African extension of the XNLI benchmark that provides natural language inference data to 15 African languages~\citep{adelani-etal-2025-irokobench}. By including AfriXNLI, we expand language coverage for the pair classification family beyond a single African language (Swahili), ensuring broader representation of African languages in entailment-style tasks.

\paragraph{EmotionAnalysisPlus}
A multi-label emotion data set that covers 32 languages, including 14 African languages~\citep{belay-etal-2025-evaluating,muhammad2025brighterbridginggaphumanannotated}. Each sentence may be assigned multiple emotion labels such as \textsc{joy}, \textsc{anger}, \textsc{sadness}, or \textsc{fear}. By including this dataset, AfriMTEB introduces the first \emph{multi-label classification} task for African languages, thereby broadening the taxonomy beyond single-label settings.

\paragraph{AfriHate}
A multilingual hate-speech classification dataset covering 14 African languages~\citep{muhammad2025afrihatemultilingualcollectionhate}. Each instance is labeled as \textsc{hate}, \textsc{abusive}, or \textsc{neutral}, providing a standardized benchmark for toxic content detection across diverse languages and registers. This dataset extends evaluation to socially relevant safety applications.

\paragraph{InjongoIntent}
A multilingual dataset for intent detection covering 16 African languages~\citep{yu2025injongomulticulturalintentdetection}. It consists of short, conversational utterances annotated with 40 everyday intent categories, for example, requests such as ``freeze account'' or ``play music.'' By focusing on dialogue-style classification, Injongo complements existing benchmarks like MASSIVE~\cite{fitzgerald-etal-2023-massive}, but offers broader African language coverage.

\paragraph{KinNews}
A Kinyarwanda news topic classification dataset with labels covering domains such as politics, business, and sports~\citep{niyongabo2020kinnewskirnewsbenchmarkingcrosslingual}. We include this dataset because \textit{MasakhaNEWS} does not cover Kinyarwanda. Adding \textit{KinNews} ensures we can cover Kinyarwanda in the AfriMTEB-Lite, since the Lite version requires maximally overlapping tasks, which is explained in the next section.  

\paragraph{\texorpdfstring{SIB200\_14Classes}{SIB200\_14Classes}}
A more challenging variant of the SIB-200 dataset, where labels are consolidated into 14 categories. This version still includes 56 African languages but is not limited to them~\citep{adelani2024sib200simpleinclusivebig}. By merging fine-grained topics into broader classes, intra-class diversity increases, which raises task difficulty. The inclusion of this dataset not only strengthens African language evaluation but also increases the difficulty for other covered languages in SIB-200, leading to more robust cross-lingual assessment.




\subsection{AfriMTEB-Lite Construction}
\label{subsec:afrimteb-lite}


The \emph{Lite} suite is introduced to address a fundamental limitation of both MMTEB and AfriMTEB-Full: \emph{uneven task--language coverage}. In large multilingual benchmarks, different tasks often include vastly different sets of languages, making macro-averaged comparisons sensitive to missing language--task pairs and inflating variance. AfriMTEB-Lite explicitly resolves this issue by enforcing uniform coverage of the same nine African languages across all tasks, enabling controlled comparisons, tighter confidence intervals, and reproducible ablation studies. The Lite suite focuses on nine geographically and typologically diverse African languages, \textit{Amharic, Oromo, Igbo, Yoruba, Hausa, Swahili, Kinyarwanda, Xhosa, and Zulu}. We retain only those datasets for which all nine languages are available, resulting in a compact yet representative benchmark of 13 datasets spanning classification, retrieval, bitext mining, clustering, pair classification, and multi-label classification.  Specifically, it comprises AfriHate, AfriSenti, AfriXNLI, Belebele retrieval, EmotionAnalysisPlus, Flores bitext mining, NTREX bitext mining, InjongoIntent, MasakhaNEWS (for seven languages) and KinNews (for Kinyarwanda) for news classification\footnote{Zulu is not covered in the news classification.}, SIB200Classification, SIB200\_14Classes, and SIB200ClusteringFast.

\begin{table*}[ht]
    \centering
    \footnotesize
    \setlength{\tabcolsep}{4pt}
        \begin{tabular}{@{}lccccccccc@{}}
\toprule
\textbf{Model} & \textbf{Btxt} & \textbf{Clf} & \textbf{Clust} & \textbf{Multi. Clf} & \textbf{Pr Clf} & \textbf{Rrnk} & \textbf{Rtrvl} & \textbf{STS} & \textbf{Avg.} \\
\midrule
\multicolumn{10}{l}{\small\emph{Small models} ($<\!1$B)} \\
\m{mmBERT-base} & 3.0 & 48.5 & 33.1 & 24.6 & 54.1 & 6.8 & 4.9 & 38.0 & 26.6 \\
\m{KaLM}   & 49.9 & 36.3 & 46.8 & 25.9 & 60.0 & 49.3 & 52.8 & 53.4 & 46.8 \\
\m{Qwen3-Embedding 0.6B}  & 33.6 & 35.9 & 37.6 & 25.2 & 58.0 & 52.9 & 52.0 & 54.1 & 43.7 \\

\m{bge-m3}   & 70.0 & 40.0 & 47.3 & 26.8 & 66.4 & \textbf{66.8} & 70.2 & 58.7 & 55.8 \\
\m{mE5-large}  & 79.7 & 43.3 & 45.3 & 27.7 & 64.3 & 65.2 & 69.2 & 62.5 & 57.2 \\
\m{mE5-large-instruct}   & \textbf{85.8} & 49.8 & 61.9 & 28.6 & 63.8 & 61.9 & 74.1 & 64.8 & 61.3 \\
\rowcolor{Gray}
\m{AfriE5-large-instruct}  & 85.5 & 49.7 & \textbf{62.9} & 29.8 & 67.9 & 64.0 & 75.4 & 63.7 & \textbf{62.4} \\

\midrule
\multicolumn{10}{l}{\small\emph{Medium models} ($\approx 4$B)} \\
\m{Qwen3-Embedding-4B} & 42.5 & 42.9 & 39.6 & 25.8 & 57.4 & 60.6 & 61.5 & 54.9 & 48.2 \\
\midrule
\multicolumn{10}{l}{\small\emph{Large models} ($\ge\!7$B)} \\
\m{gte-Qwen2-7B-instruct}  & 52.7 & 41.0 & 56.6 & 25.1 & 58.1 & 58.8 & 60.3 & 54.9 & 50.9 \\
\m{GritLM-7B}   & 45.2 & 43.4 & 54.0 & 26.6 & 59.6 & 65.4 & 61.0 & 59.1 & 51.8 \\
\m{Linq-Embed-Mistral} & 45.2 & 43.2 & 55.4 & 27.1 & 59.4 & 65.1 & 59.2 & 62.5 & 52.1 \\
\m{SFR-Embedding-Mistral} & 46.3 & 42.5 & 58.0 & 26.2 & 58.9 & 63.8 & 58.1 & 62.0 & 52.0 \\
\m{E5 mistral 7b instruct}  & 46.4 & 41.9 & 58.5 & 26.0 & 58.7 & 64.1 & 57.3 & 61.3 & 51.8 \\
\m{Qwen3-Embedding-8B} & 48.4 & 43.7 & 41.8 & 27.2 & 58.3 & 60.0 & 69.9 & 54.7 & 50.5 \\
\midrule
\multicolumn{10}{l}{\small\emph{Undisclosed size}} \\
\m{gemini embedding 001}  & 72.2 & \textbf{50.0} & 52.7 & \textbf{32.7} & \textbf{71.6} & 63.4 & \textbf{77.5} & \textbf{65.0} & 60.6 \\
\bottomrule
\end{tabular}
   \vspace{-2mm}
    \caption{\textbf{AfriMTEB-Full results.} Average performance of embedding models across languages grouped by task family. 
    The final column reports the unweighted macro-average across task families. Best scores in each column are highlighted in bold. Differences between AfriE5 and mE5 that are statistically significant under paired bootstrap testing are reported in Appendix~\ref{app:statistical_significance}.}
    \label{tab:multi_lang_afrimteb}
    \vspace{-2mm}
\end{table*}

\section{Adapting Embedding Models to African Languages}

\begin{table*}[ht]
  \centering
  \footnotesize
  \setlength{\tabcolsep}{4pt}
  \resizebox{\textwidth}{!}{
  \begin{tabular}{@{}l*{13}{r}@{}}
    \toprule
    & \multicolumn{3}{c}{}%
    & \multicolumn{1}{c}{} &
    & \multicolumn{2}{c}{\textbf{Btxt}} &
    & \multicolumn{1}{c}{}%
    & \multicolumn{3}{c}{\textbf{SIB-200}}%
    & \multicolumn{1}{c}{}\\
    \cmidrule(lr){7-8}
    \cmidrule(lr){11-13}
    \textbf{Model} &
    \textbf{Hate} & \textbf{Senti} & \textbf{NLI} &
    \textbf{Retrvl} & \textbf{Emo} &
    \textbf{Flores} & \textbf{NTREX} &
    \textbf{Intent} &
    \textbf{News} &
    \textbf{14Classes} & \textbf{Class} & \textbf{Clust} &
    \textbf{Avg.} \\
    \midrule
    \multicolumn{14}{l}{\small\emph{Small models} ($<\!1$B)} \\

    \m{mmBERT-base} &
    48.2 & 39.9 & 54.0 &
    4.0 & 27.3 &
    1.8 & 2.2 &
    72.5 &
    55.1 &
    2.4 & 34.2 & 5.0 &
    28.9 \\

    \m{KaLM} &
    48.5 & 44.3 & 61.0 &
    51.0 & 28.9 &
    53.2 & 58.6 &
    63.4 &
    74.6 &
    7.3 & 48.9 & 16.8 &
    46.4 \\

    \m{Qwen3-Embedding 0.6B} &
    48.6 & 41.5 & 58.1 &
    46.3 & 28.1 &
    33.0 & 38.1 &
    68.5 &
    70.5 &
    3.5 & 40.0 & 12.0 &
    40.7 \\

    \m{EmbeddingGemma 300m} &
    45.4 & 39.0 & 53.6 &
    10.6 & 26.6 &
    12.2 & 17.9 &
    49.4 &
    59.0 &
    1.4 & 29.6 & 4.6 &
    29.1 \\

    \m{bge-m3} &
    50.1 & 47.9 & 68.7 &
    69.9 & 29.4 &
    78.1 & 80.4 &
    75.4 &
    72.7 &
    10.2 & 55.6 & 21.0 &
    55.0 \\

    \m{mE5-large} &
    49.8 & 48.9 & 65.2 &
    69.9 & 31.3 &
    86.9 & 88.7 &
    77.1 &
    77.7 &
    11.6 & 60.4 & 25.6 &
    57.8 \\

    \m{mE5-large-instruct} &
    51.5 & 47.0 & 64.5 &
    75.7 & 31.5 &
    \textbf{91.4} & 91.5 &
    75.5 &
    78.8 &
    22.0 & 71.2 & 43.9 &
    62.0 \\

    \rowcolor{Gray}
    \m{AfriE5-large-instruct} &
    51.7 & 50.7 & 69.0 &
    77.7 & 32.8 &
    91.2 & \textbf{92.0} &
    75.4 &
    \textbf{79.5} &
    \textbf{26.2} & \textbf{72.0} & \textbf{45.7} &
    \textbf{63.7} \\

    \midrule
    \multicolumn{14}{l}{\small\emph{Medium models} ($\approx 4$B)} \\

    \m{Qwen3-Embedding-4B} &
    50.0 & 41.2 & 56.8 &
    58.3 & 28.3 &
    47.7 & 49.0 &
    70.1 &
    76.0 &
    7.9 & 49.2 & 17.5 &
    46.0 \\

    \midrule
    \multicolumn{14}{l}{\small\emph{Large models} ($\ge\!7$B)} \\

    \m{gte-Qwen2-7B-instruct} &
    46.0 & 43.5 & 58.6 &
    55.6 & 27.8 &
    58.4 & 60.0 &
    57.1 &
    79.8 &
    10.4 & 50.8 & 22.4 &
    47.5 \\

    \m{GritLM-7B} &
    51.4 & 45.6 & 59.8 &
    55.0 & 29.6 &
    46.5 & 52.0 &
    70.8 &
    75.3 &
    8.8 & 50.8 & 22.2 &
    47.3 \\

    \m{Linq-Embed-Mistral} &
    51.0 & 43.8 & 59.7 &
    52.1 & 30.0 &
    46.7 & 52.0 &
    70.3 &
    76.5 &
    7.6 & 50.7 & 22.1 &
    46.9 \\

    \m{SFR-Embedding-Mistral} &
    49.3 & 44.5 & 58.9 &
    50.9 & 28.8 &
    47.7 & 53.0 &
    62.5 &
    77.0 &
    8.5 & 49.7 & 21.9 &
    46.1 \\

    \m{E5 mistral 7b instruct} &
    49.0 & 45.7 & 58.7 &
    50.2 & 28.6 &
    47.9 & 53.3 &
    61.8 &
    76.0 &
    7.2 & 49.2 & 21.8 &
    45.8 \\

    \m{Qwen3-Embedding 8B} &
    50.8 & 46.2 & 58.8 &
    68.3 & 29.1 &
    58.3 & 57.3 &
    67.7 &
    77.6 &
    8.0 & 53.5 & 20.9 &
    49.7 \\

    \midrule
    \multicolumn{14}{l}{\small\emph{Undisclosed size}} \\

    \m{gemini embedding 001} &
    \textbf{55.0} & \textbf{53.8} & \textbf{75.3} &
    \textbf{83.6} & \textbf{35.5} &
    88.1 & 84.2 &
    \textbf{83.9} &
    76.8 &
    17.7 & 69.2 & 34.6 &
    63.1 \\

    \bottomrule
  \end{tabular}}
  \vspace{-2mm}
  \caption{\textbf{AfriMTEB-Lite results.} Average performance of embedding models across nine African languages (\texttt{AMH, GAZ, HAU, IBO, KIN, SWA, XHO, YOR, ZUL}) on 12 tasks. The final column gives the unweighted macro average across tasks. Best scores per column are highlighted in bold. Differences between AfriE5 and mE5 that are statistically significant under paired bootstrap testing are reported in Appendix~\ref{app:statistical_significance}.}
  \label{tab:afromteb_dataset_results}
  \vspace{-3mm}
\end{table*}

While AfriMTEB provides a standardized and task-diverse evaluation framework for African languages, it also exposes clear performance gaps in existing multilingual embedding models. To demonstrate how such gaps can be addressed in a data-efficient manner, we next present \textbf{\afriefive}, an embedding model adapted using cross-lingual contrastive distillation and evaluated systematically under the AfriMTEB.

\subsection{Method}

Our approach combines contrastive learning with knowledge distillation in a unified objective. Given a batch of $B$ queries, each associated with a group of $G$ passages, the total loss is:
\vspace{-2mm}
\begin{equation}
\mathcal{L} = \mathcal{L}_{contrastive} + \mathcal{L}_{kd}.
\end{equation}



\paragraph{Contrastive Learning}
The contrastive learning term is computed as follows:
\begin{equation}
\begin{split}
\mathcal{L}_{\text{contrastive}}
  = -\frac{1}{B} \sum_{i=1}^{B} 
  \log \frac{\exp(s_{i,pos} / \tau)}
           {\sum_{j=1}^{B \cdot G} \exp(s_{i,j} / \tau)}.
\end{split}
\end{equation}

Here $s_{i,j} = \text{cos}(q_i, p_j)$ is the similarity between the query embedding $\mathbf{q}_i$ and passage embedding $\mathbf{p}_j$, and $\tau$ is a temperature hyperparameter. The numerator contains the similarity between query $\mathbf{q}_i$ and its corresponding positive passage $\mathbf{p}_\text{pos}$. The denominator aggregates similarities between $\mathbf{q}_i$ and all passages in the batch, covering both pre-mined hard negatives (derived from NLI contradiction examples and hard negative mining) and in-batch negatives (passages associated with other queries in the same training batch). 

\paragraph{Knowledge Distillation}  
The distillation term aligns the student’s predicted distribution with the teacher’s scores using cross-entropy:

\vspace{-3mm}
\begin{equation}
\mathcal{L}_{kd} = -\frac{1}{B} \sum_{i=1}^{B} \sum_{j=1}^{G} P_{teacher}^{(i,j)} \log P_{student}^{(i,j)},
\end{equation}
\vspace{-2mm}

where $P_{teacher}^{(i,j)}$ and $P_{student}^{(i,j)}$ are normalized softmax scores for the query $i$ and the $j$-th passage, produced by the teacher reranker and the student encoder, respectively.\footnote{We used embedding cosine similarity for the student encoder and logits of the concatenated query and passage for the teacher reranker.} Teacher's scores are obtained during training data construction, where each example is annotated with both positive/negative labels and teacher's score. We detail this data creation process in the next section.

\subsection{Cross-lingual Training Data Construction}
\label{sec:cross_lingual}

We constructed cross-lingual training data by leveraging large-scale natural language inference corpora, a supervision signal that has proven effective for learning sentence embeddings~  \citep{gao2022simcsesimplecontrastivelearning,reimers2019sentencebertsentenceembeddingsusing}. Specifically, we used MultiNLI and SNLI \citep{williams2018broadcoveragechallengecorpussentence,bowman2015largeannotatedcorpuslearning} as source datasets in English. Each sentence pair was translated into the nine African target languages using NLLB-200 (3.3B) \citep{nllbteam2022languageleftbehindscaling}. We then estimated translation quality using SSA-COMET-MTL~\citep{li2025ssacometllmsoutperformlearned}, a COMET variant~\cite{rei-etal-2020-comet} that covers the target African languages, and filtered pairs below a threshold of 0.75 to ensure data quality. We select SSA-COMET because it currently shows the strongest correlation with human judgments for African language pairs among available MT quality estimation metrics.\footnote{We analyze the effect of different quality thresholds on the resulting sample size in Appendix~\ref{sec:ssa_comet_samples}.}

To encourage cross-lingual alignment, each example was expanded into multiple configurations: (i) premise in the target language and hypothesis in the source, (ii) premise in the source and hypothesis in the target, (iii) both in the target language, and (iv) both in the source language ( i.e., English). For MultiNLI, we reformulated examples as query–positive/negative pairs (entailment as \textit{pos}, contradiction as \textit{neg}). For SNLI, we followed the same strategy but included all two-way annotations (positive and negative). 
We employ hard negative mining using \textit{mE5-Large-Instruct} \citep{wang2024multilinguale5textembeddings} to identify challenging negative examples. For each query, we encode all corpus passages and perform FAISS-based nearest neighbor search to retrieve the top-$k$ most similar passages. We sample 15 hard negatives from a rank window (ranks 2-200), excluding positives and the query itself. Hard negatives are semantically similar to the query but irrelevant, forcing the model to learn fine-grained discrimination. 

\subsection{Experimental Settings}

We fine-tuned \textit{mE5-Large-Instruct}~\citep{wang2024multilinguale5textembeddings} on our curated cross-lingual training data using the open-sourced \texttt{FlagEmbedding}\footnote{\url{https://github.com/FlagOpen/FlagEmbedding}} training repository. We followed the default implementation of contrastive learning with cross-device negatives and enabled knowledge distillation with teacher scores provided by BGE Reranker v2 m3~\citep{chen2024bgem3embeddingmultilingualmultifunctionality}. All samples in a batch were drawn from the same dataset to maintain consistent supervision. Training was performed on a single GPU and the key hyperparameters are summarized in Table~\ref{tab:train-config}. We trained the model for one epoch with logging every 100 steps and checkpoint saving every 100 steps. The resulting model is referred to as \textbf{\afriefive}. We provide brief descriptions of the baseline models in Appendix \ref{subsec:baseline}.

\section{Results}










\subsection{AfriMTEB Results}

\paragraph{Smaller-sized E5 variants have comparable performance to Gemini Embedding.} As shown in \autoref{tab:multi_lang_afrimteb},
despite being small models ($<1$B), the mE5 family matches or surpasses the proprietary model, gemini embedding 001, especially on bitext mining (i.e., Btxt). \afriefive{} attains the best macro average at 62.4, edging out \textit{Gemini embedding} (60.6) and \mefive{} (61.3). This indicates that strong multilingual coverage and targeted adaptation can outweigh model size or access to proprietary training data. Paired bootstrap tests in Appendix~\ref{app:statistical_significance} show that the overall macro improvement of \afriefive{} over \mefive{} is statistically significant on AfriMTEB-Full.

\paragraph{AfriE5 and mE5 excel on bitext mining and clustering.}
On bitext mining, \mefive{} and \afriefive{} have comparable performance of $85$ points, both are far ahead of other opens and the API baseline (\textit{Gemini embedding} at 72.2). For clustering, \afriefive{} leads with 62.9 and \mefive{} is next (61.9), while larger models and Gemini embeddings are behind.
These families reward models that learn language-agnostic semantic spaces with robust cross-lingual alignment, which appears to be a particular strength of the E5 lineage.

\paragraph{AfriE5 improves reranking with cross-lingual training dataset and knowledge distillation.}
Although the base model \mefive{} records a reranking score of 61.9, \afriefive{} raises this to 64.0. It also surpasses Gemini embedding 001 (63.4) on the same task (Table~\ref{tab:multi_lang_afrimteb}). This improvement can be attributed to the training recipe, where \afriefive{} leverages knowledge distillation from the \textsc{BGE-m3} cross-encoder in addition to contrastive supervision. The result suggests that incorporating reranker-derived soft labels enhances cross-lingual alignment for ranking-style objectives.

\paragraph{Gemini Embedding excels on classification tasks over E5 variants.}
While E5 variants lead overall, \textit{Gemini embedding 001} is strongest on most classification-style families: single-label classification (50.0 vs.\ 49.8 for \mefive{} and 49.7 for \afriefive), multi-label classification (32.7 vs.\ 28.6/29.8), and pair classification (71.6 vs.\ 63.8/67.9). It also tops retrieval (77.5) and semantic textual similarity (65.0). These strengths suggest Gemini’s instruction and data mix particularly benefits discriminative judgment tasks and sentence-level similarity scoring.

\paragraph{Language coverage is more crucial than text embedding model sizes.}
Large 7B and 8B encoders do not translate into higher \afrimteb{} scores. All 7B/8B model scores cluster in the low–mid 50s, for example, \textit{gte-Qwen2-7B-instruct} at 50.9, \textit{GritLM-7B} at 51.8, \textit{Qwen3-Embedding-8B} at 50.5. 
They are clearly below the smaller E5 variants of around 61.3 point. This gap underscores that broad, balanced language coverage and task diversity matter more than parameter count alone.

\paragraph{AfriE5 generalizes to 59 languages despite training on only 9.}
Although \afriefive{} is adapted using supervision centered on only nine African languages, it achieves the highest macro average (62.4) across 59 languages. Relative to \mefive{}, \afriefive{} shows consistent family-level gains on pair classification (+4.1), reranking (+2.1), retrieval (+1.3), clustering (+1.0), and multi-label classification (+1.2), with small trade-offs on bitext mining (–0.3), single-label classification (–0.1), and STS (–1.1). Importantly, paired bootstrap analysis over task--language cells shows that improvements in pair classification, reranking, retrieval, and multi-label classification are statistically significant, while changes in other families are not (Appendix~\ref{app:statistical_significance}). This pattern indicates that targeted cross-lingual distillation yields transferable gains beyond the training languages, particularly for retrieval-style and alignment-sensitive tasks.

\begin{figure*}[h]
    \centering
    \includegraphics[width=\linewidth]{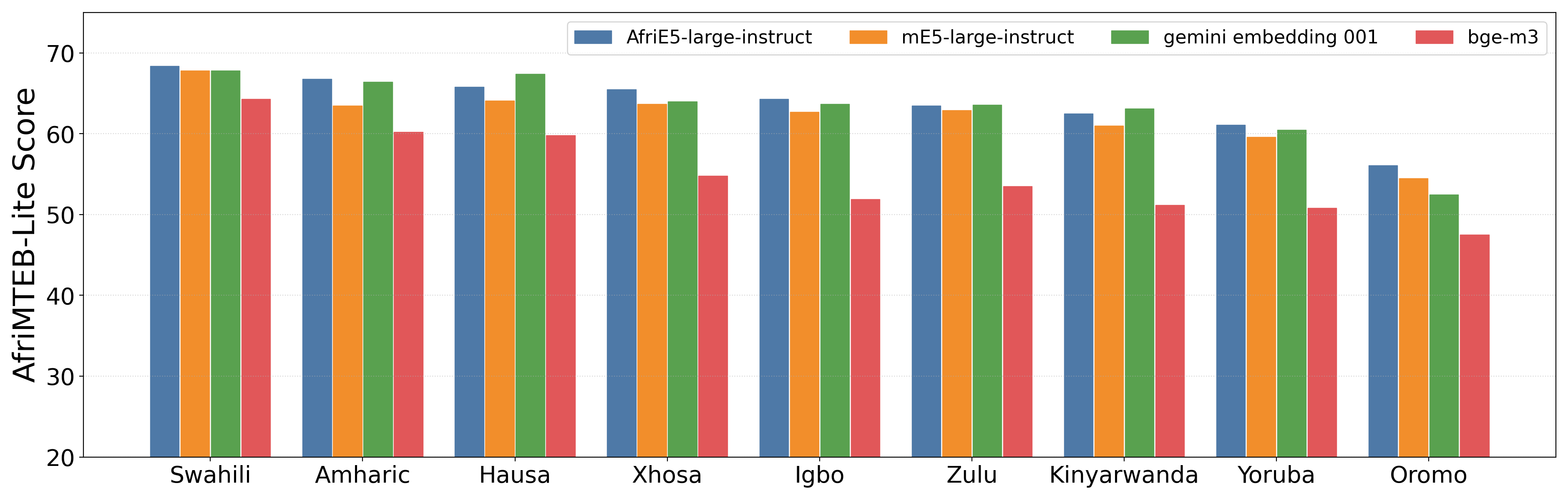}
    \caption{Performance on \textbf{AfriMTEB-Lite} across nine target languages. Bars show average scores for four representative embedding models. AfriE5-large-instruct consistently achieves the highest or near-highest scores.}
    \label{fig:afrimteb-lite-results}
    \vspace{-3mm}
\end{figure*}


\subsection{AfriMTEB-Lite Results}
Here, we present the overall aggregated results by dataset and language in \autoref{tab:afromteb_dataset_results} and \autoref{fig:afrimteb-lite-results}. The comprehensive results are in Appendix~\ref{sec:detailed_afrimteblite}. 

\paragraph{AfriE5 achieves the strongest overall performance on AfriMTEB-Lite.}
On the Lite suite of nine African languages and twelve tasks, \afriefive{} attains the highest overall macro average with a score of 63.7, compared to 62.0 for \mefive{} and 63.1 for \textit{Gemini embedding} (Table~\ref{tab:afromteb_dataset_results}). While performance differences vary across task families, paired bootstrap tests over languages show statistically significant improvements for \afriefive{} on 9 of the 12 tasks, including AfriXNLI, SIB200 variants, clustering, and retrieval, as well as for the overall macro- and micro-averages (Appendix~\ref{app:statistical_significance}). In addition, multi-seed experiments confirm that these gains are stable, with standard deviation within $\pm 0.1$ on the Lite macro score (Appendix~\ref{app:multiple_seeds}). These results demonstrate that cross-lingual contrastive distillation improves effectiveness for African languages even when training is restricted to a compact set of nine languages.


\paragraph{AfriE5 achieves the highest macro averages on 6 languages out of 9.}
Figure~\ref{fig:afrimteb-lite-results} summarizes language-level macro averages across all Lite tasks, with exact values reported in Table~\ref{tab:afromteb_dataset_results}. AfriE5 achieves the highest average performance on six of the nine languages—Swahili, Amharic, Xhosa, Igbo, Yoruba, and Oromo; consistently outperforming both \mefive{} and \textit{Gemini embedding 001}. The gains are particularly pronounced for lower-resource languages such as Oromo and Xhosa, where AfriE5 shows clear margins over all baselines.

\paragraph{Our adaptation boosts classification tasks, especially SIB200\_14Classes.}
Relative to \mefive{}, \afriefive{} shows clear improvements on several classification benchmarks. It achieves higher scores on AfriSenti (+3.7), 
AfriXNLI (+4.5), 
and SIB200clustering (+1.8). 
The largest gain appears on SIB200\_14Classes (+4.4; 26.2 vs.\ 21.8), a setting that requires grouping semantically diverse texts into broad topical categories. These improvements indicate that the cross-lingual, NLI-style contrastive learning used in AfriE5 strengthens semantic alignment across languages, leading to more effective cross-lingual transfer for classification and clustering tasks.

\begin{table*}[ht]
    \centering
    \footnotesize
    \setlength{\tabcolsep}{4pt}
    \resizebox{\textwidth}{!}{
    \begin{tabular}{c c|*{11}{r}|r}
    \toprule
    & & & & & 
    & \multicolumn{2}{c}{\textbf{Btxt}} 
    & & & & \textbf{SIB-200} &\\
    \cmidrule(lr){7-8}
    \cmidrule(lr){11-13}
    \textbf{Cross-lingual Expansion} & \textbf{QE Thres.} & \textbf{Hate} & \textbf{Senti} & \textbf{NLI} & \textbf{Emo} & \textbf{Flores} & \textbf{NTREX} & \textbf{Intent} & \textbf{News}  & \textbf{14Classes} & \textbf{Class} & \textbf{Clust} & \textbf{Avg.} \\
    \midrule
    \cmark & 0.67 & \textbf{52.2} & 50.9 & 66.3 & 32.3 & \textbf{91.6} & \textbf{94.1}  & \textbf{77.6} & \textbf{82.7} & 23.8 & 71.7 & 44.8 & 62.5 \\
    \cmark & 0.70 & 51.3 & 52.0 & 69.0 & 32.6 & 91.3 & 92.2 & 75.2 & 79.5 & 26.9 & 73.1 & 45.5 & 62.6 \\
    \rowcolor{Gray}
    \cmark & 0.75 & 51.4 & \textbf{53.3} & \textbf{69.1} & \textbf{32.8} & 91.2 & 93.8 & 77.1 & 82.6  & \textbf{26.2} & \textbf{72.0} & \textbf{45.7} & \textbf{63.2} \\
    \cmark & 0.80 & 51.5 & 49.9 & 68.2 & 29.8 & 88.2 & 91.0 & 85.9 & 79.9  & 11.2 & 61.8 & 24.2 & 58.3 \\
    \midrule
    \rowcolor{Gray}
    \xmark & 0.75 & 52.3 & 52.8 & 68.2 & 32.0 & 88.5 & 91.2 & 81.2 & 82.0 & 21.8 & 70.6 & 45.2 & 62.3 \\
    
    \bottomrule
    \end{tabular}
    }
    \caption{\textbf{Ablation study of AfriE5-large-instruct on AfriMTEB-Lite.} We vary the use of cross-lingual dataset expansion and the translation quality threshold during filtering. Best scores per column are highlighted in bold. Note that Belebele retrieval is not covered in this ablation study.}
    \label{tab:ablation_lite}
\end{table*}

\section{Ablation}

We conduct controlled experiments where all training settings are held fixed (base architecture, loss, knowledge distillation from BGE reranker, batch/group sizes, number of steps, and sampling) while varying a single factor at a time. The results are reported in Table~\ref{tab:ablation_lite}.

\paragraph{Impact of cross-lingual dataset expansion}  
As described in Section~\S\ref{sec:cross_lingual}, we compare training setups that differ only in whether cross-lingual dataset expansion is applied. 
Without cross-lingual dataset expansion (\xmark), each NLI example is used within a single language at a time; only target--target sentence pairs (e.g., Swahili--Swahili) are constructed.
With cross-lingual dataset expansion (\cmark), each example is expanded into four configurations: target--target, English--English, target--English, and English--target.
Comparing row 3 (\xmark, 0.75) and row 5 (\cmark, 0.75), the average increases from 62.3 to 63.2. The cross-lingual expansion of the data set exposes the model to richer cross-lingual contrasts: First, higher scores on bitext mining tasks show that the expansion improves the model’s ability to align semantically equivalent sentences between languages, producing a more coherent multilingual embedding space. Second, classification tasks such as SIB200-14 benefit from this stronger alignment, as the model learns to abstract over diverse topical domains without relying on language-specific cues. \footnote{Belebele retrieval was added to AfriMTEB-Lite after the ablation study had been finalized; therefore, results on Belebele are not reported in Table 4.}

\paragraph{Machine translation quality threshold (QE)}  
With expansion enabled (rows 1–4), the best overall score is achieved after filtering using a COMET variant, SSA-COMET-MTL QE=0.75 (63.2), compared to 62.5 at 0.67 and a sharp drop to 58.3 at 0.80. A low threshold (0.67) retains large noisy translations (around 433K training samples), which slightly hurts reasoning-heavy tasks. In contrast, a high threshold (0.80) filters out too much data (remaining only 7500 samples), reducing linguistic and semantic diversity and leading to weaker transfer, especially in classification. The middle ground (0.75) balances translation quality with coverage with around 60K samples, yielding the most reliable improvements across tasks.  We provide the number of samples retained by SSA-COMET-MTL in Appendix~\ref{sec:ssa_comet_samples}.

\section{Related Work}

\label{sec:related}

\paragraph{Multilingual text embedding benchmarks}
The Massive Text Embedding Benchmark (MTEB) introduced a common taxonomy and leaderboard for sentence embeddings \citep{muennighoff2023mtebmassivetextembedding}. Since then, several language- or region-specific variants have emerged, including MMTEB (with Europe and Indic tracks), SEA-BED for Southeast Asia, PL-MTEB for Polish, and MTEB-French \citep{ponwitayarat2025seabedsoutheastasiaembedding,poświata2024plmtebpolishmassivetext,ciancone2024mtebfrenchresourcesfrenchsentence}. Additional efforts target specific language families or regions: C-MTEB (Chinese), German-focused suites, JMTEB (Japanese), Korean-focused suites, ruMTEB (Russian), FaMTEB (Persian/Farsi), and VN-MTEB (Vietnamese) \citep{xiao2024cpackpackedresourcesgeneral,wehrli2024germantextembeddingclustering,jmteb,snegirev2025russianfocusedembeddersexplorationrumteb,zinvandi2025famtebmassivetextembedding,pham2025vnmtebvietnamesemassivetext}. Our work follows this trend by building an Africa-focused extension with broad task coverage.

\paragraph{Multilingual text embedding models}
Commercial/API models widely used in practice include OpenAI’s text-embedding-3 series, Google’s gemini-embedding-001, and Cohere’s Embed v3 \citep{neelakantan2022textcodeembeddingscontrastive,lee2025geminiembeddinggeneralizableembeddings}. Open-weight models are an active research area: me5, e5, and e5-mistral-7b-instruct provide strong general-purpose and instruction-tuned baselines \citep{wang2024multilinguale5textembeddings,wang2024textembeddingsweaklysupervisedcontrastive}; Qwen3-Embedding and Embedding Gemma offer lightweight multilingual options \citep{zhang2025qwen3embeddingadvancingtext,vera2025embeddinggemmapowerfullightweighttext}; GTE proposes efficient, general-purpose embeddings \citep{li2023generaltextembeddingsmultistage}; and classic multilingual encoders LaBSE remain strong references \citep{feng2022languageagnosticbertsentenceembedding}. The recent BGE-M3 model integrates multilingual, multi-function training and is a competitive open baseline \citep{chen2024bgem3embeddingmultilingualmultifunctionality}. Despite progress, coverage and performance on many African languages remain uneven, motivating region-specific evaluation and targeted adaptation.

\section{Conclusion}

We presented AfriMTEB, a large-scale benchmark for African languages spanning 59 languages and 14 tasks, and AfriE5, an adaptation of mE5-large-instruct via cross-lingual contrastive distillation. AfriE5 achieves the strongest overall macro-average among open-weight embedding models on both AfriMTEB-Full and AfriMTEB-Lite, with statistically significant gains on several task families, while remaining competitive with proprietary baselines such as Gemini Embedding-001. Our ablations show that cross-lingual dataset expansion and balanced translation filtering are crucial for these gains. Together, AfriMTEB and AfriE5 provide a standardized evaluation framework and a practical reference point for advancing text embedding research for African languages.

\section*{Limitations}

AfriMTEB expands coverage to 59 languages, yet many African languages, dialects, orthographies, and code-switched registers remain under-represented. Several datasets inherit noise or heterogeneity from crowd labels, repurposed tasks, and preprocessing. Moreover, our adaptation data relies on machine translation via NLLB-200 and automatic quality estimation using SSA-COMET, which vary in reliability across language pairs and domains. Distillation from a single teacher (BGE Reranker v2 m3) can also transfer its biases, potentially advantaging languages and styles that align with the MT/QE pipeline and the teacher’s preferences.

Our evaluation is limited to text-only sentence and paragraph embeddings and reports macro-averages across tasks and languages; alternative weightings (e.g., by population or application criticality) and additional metrics (calibration, robustness, fairness) could yield different conclusions. End-to-end RAG quality, multimodal retrieval, and very long-context document embeddings are out of scope, and domain coverage skews toward formal/news text over colloquial or specialized domains. True parity with closed-weight baselines (e.g., data mixtures, architectures, inference settings) is infeasible; repeated API evaluations may be affected by nondeterminism, and some open models may be sensitive to prompt formats or poolers we did not exhaustively tune.

While AfriE5 trained on nine languages generalizes well to 59 in our tests, transfer may degrade for typologically distant or extremely low-resource languages; broader community datasets and multi-teacher/multi-signal training are promising avenues to mitigate these limitations.

\section*{Acknowledgements}
This research was supported in part by the Natural Sciences and Engineering Research Council (NSERC) of Canada. David Adelani acknowledges the funding of IVADO and the Canada First Research Excellence Fund. We acknowledge the use of Gen AI tools for grammar checking, and no scientific content was generated by the tools.


\bibliography{custom}

@inproceedings{muennighoff2023mtebmassivetextembedding,
    title = "{MTEB}: Massive Text Embedding Benchmark",
    author = "Muennighoff, Niklas  and
      Tazi, Nouamane  and
      Magne, Loic  and
      Reimers, Nils",
    editor = "Vlachos, Andreas  and
      Augenstein, Isabelle",
    booktitle = "Proceedings of the 17th Conference of the European Chapter of the Association for Computational Linguistics",
    month = may,
    year = "2023",
    address = "Dubrovnik, Croatia",
    publisher = "Association for Computational Linguistics",
    url = "https://aclanthology.org/2023.eacl-main.148/",
    doi = "10.18653/v1/2023.eacl-main.148",
    pages = "2014--2037"
}

@misc{gao2022simcsesimplecontrastivelearning,
      title={SimCSE: Simple Contrastive Learning of Sentence Embeddings}, 
      author={Tianyu Gao and Xingcheng Yao and Danqi Chen},
      year={2022},
      eprint={2104.08821},
      archivePrefix={arXiv},
      primaryClass={cs.CL},
      url={https://arxiv.org/abs/2104.08821}, 
}

@misc{feng2022languageagnosticbertsentenceembedding,
      title={Language-agnostic BERT Sentence Embedding}, 
      author={Fangxiaoyu Feng and Yinfei Yang and Daniel Cer and Naveen Arivazhagan and Wei Wang},
      year={2022},
      eprint={2007.01852},
      archivePrefix={arXiv},
      primaryClass={cs.CL},
      url={https://arxiv.org/abs/2007.01852}, 
}

@misc{adelani2024sib200simpleinclusivebig,
      title={SIB-200: A Simple, Inclusive, and Big Evaluation Dataset for Topic Classification in 200+ Languages and Dialects}, 
      author={David Ifeoluwa Adelani and Hannah Liu and Xiaoyu Shen and Nikita Vassilyev and Jesujoba O. Alabi and Yanke Mao and Haonan Gao and Annie En-Shiun Lee},
      year={2024},
      eprint={2309.07445},
      archivePrefix={arXiv},
      primaryClass={cs.CL},
      url={https://arxiv.org/abs/2309.07445}, 
}

@misc{wang2024textembeddingsweaklysupervisedcontrastive,
      title={Text Embeddings by Weakly-Supervised Contrastive Pre-training}, 
      author={Liang Wang and Nan Yang and Xiaolong Huang and Binxing Jiao and Linjun Yang and Daxin Jiang and Rangan Majumder and Furu Wei},
      year={2024},
      eprint={2212.03533},
      archivePrefix={arXiv},
      primaryClass={cs.CL},
      url={https://arxiv.org/abs/2212.03533}, 
}

@misc{wang2024multilinguale5textembeddings,
      title={Multilingual E5 Text Embeddings: A Technical Report}, 
      author={Liang Wang and Nan Yang and Xiaolong Huang and Linjun Yang and Rangan Majumder and Furu Wei},
      year={2024},
      eprint={2402.05672},
      archivePrefix={arXiv},
      primaryClass={cs.CL},
      url={https://arxiv.org/abs/2402.05672}, 
}

@misc{muhammad2025semeval2025task11bridging,
      title={SemEval-2025 Task 11: Bridging the Gap in Text-Based Emotion Detection}, 
      author={Shamsuddeen Hassan Muhammad and Nedjma Ousidhoum and Idris Abdulmumin and Seid Muhie Yimam and Jan Philip Wahle and Terry Ruas and Meriem Beloucif and Christine De Kock and Tadesse Destaw Belay and Ibrahim Said Ahmad and Nirmal Surange and Daniela Teodorescu and David Ifeoluwa Adelani and Alham Fikri Aji and Felermino Ali and Vladimir Araujo and Abinew Ali Ayele and Oana Ignat and Alexander Panchenko and Yi Zhou and Saif M. Mohammad},
      year={2025},
      eprint={2503.07269},
      archivePrefix={arXiv},
      primaryClass={cs.CL},
      url={https://arxiv.org/abs/2503.07269}, 
}

@misc{zhang2021mrtydimultilingualbenchmark,
      title={Mr. TyDi: A Multi-lingual Benchmark for Dense Retrieval}, 
      author={Xinyu Zhang and Xueguang Ma and Peng Shi and Jimmy Lin},
      year={2021},
      eprint={2108.08787},
      archivePrefix={arXiv},
      primaryClass={cs.CL},
      url={https://arxiv.org/abs/2108.08787}, 
}

@article{goyal-etal-2022-flores,
    title = "The {F}lores-101 Evaluation Benchmark for Low-Resource and Multilingual Machine Translation",
    author = "Goyal, Naman  and
      Gao, Cynthia  and
      Chaudhary, Vishrav  and
      Chen, Peng-Jen  and
      Wenzek, Guillaume  and
      Ju, Da  and
      Krishnan, Sanjana  and
      Ranzato, Marc{'}Aurelio  and
      Guzm{\'a}n, Francisco  and
      Fan, Angela",
    editor = "Roark, Brian  and
      Nenkova, Ani",
    journal = "Transactions of the Association for Computational Linguistics",
    volume = "10",
    year = "2022",
    address = "Cambridge, MA",
    publisher = "MIT Press",
    url = "https://aclanthology.org/2022.tacl-1.30/",
    doi = "10.1162/tacl_a_00474",
    pages = "522--538"
}

@inproceedings{kann-2024-massively,
    title = "Massively Multilingual Token-Based Typology Using the Parallel {B}ible Corpus",
    author = "Kann, Amanda",
    editor = "Calzolari, Nicoletta  and
      Kan, Min-Yen  and
      Hoste, Veronique  and
      Lenci, Alessandro  and
      Sakti, Sakriani  and
      Xue, Nianwen",
    booktitle = "Proceedings of the 2024 Joint International Conference on Computational Linguistics, Language Resources and Evaluation (LREC-COLING 2024)",
    month = may,
    year = "2024",
    address = "Torino, Italia",
    publisher = "ELRA and ICCL",
    url = "https://aclanthology.org/2024.lrec-main.965/",
    pages = "11070--11079"
}

@misc{shode2023nollysentileveragingtransferlearning,
      title={NollySenti: Leveraging Transfer Learning and Machine Translation for Nigerian Movie Sentiment Classification}, 
      author={Iyanuoluwa Shode and David Ifeoluwa Adelani and Jing Peng and Anna Feldman},
      year={2023},
      eprint={2305.10971},
      archivePrefix={arXiv},
      primaryClass={cs.CL},
      url={https://arxiv.org/abs/2305.10971}, 
}

@misc{tiedemann2020tatoebatranslationchallenge,
      title={The Tatoeba Translation Challenge -- Realistic Data Sets for Low Resource and Multilingual MT}, 
      author={Jörg Tiedemann},
      year={2020},
      eprint={2010.06354},
      archivePrefix={arXiv},
      primaryClass={cs.CL},
      url={https://arxiv.org/abs/2010.06354}, 
}

@misc{conneau2018xnlievaluatingcrosslingualsentence,
      title={XNLI: Evaluating Cross-lingual Sentence Representations}, 
      author={Alexis Conneau and Guillaume Lample and Ruty Rinott and Adina Williams and Samuel R. Bowman and Holger Schwenk and Veselin Stoyanov},
      year={2018},
      eprint={1809.05053},
      archivePrefix={arXiv},
      primaryClass={cs.CL},
      url={https://arxiv.org/abs/1809.05053}, 
}

@misc{madodonga2023izindabatindzabamachinelearningnews,
      title={Izindaba-Tindzaba: Machine learning news categorisation for Long and Short Text for isiZulu and Siswati}, 
      author={Andani Madodonga and Vukosi Marivate and Matthew Adendorff},
      year={2023},
      eprint={2306.07426},
      archivePrefix={arXiv},
      primaryClass={cs.CL},
      url={https://arxiv.org/abs/2306.07426}, 
}

@misc{south-african-language-identification,
    author = {ExploreAI Academy and Joanne M},
    title = {South African Language Identification},
    year = {2022},
    howpublished = {\url{https://kaggle.com/competitions/south-african-language-identification}},
    note = {Kaggle}
}

@misc{ousidhoum2024semrel2024collectionsemantictextual,
      title={SemRel2024: A Collection of Semantic Textual Relatedness Datasets for 13 Languages}, 
      author={Nedjma Ousidhoum and Shamsuddeen Hassan Muhammad and Mohamed Abdalla and Idris Abdulmumin and Ibrahim Said Ahmad and Sanchit Ahuja and Alham Fikri Aji and Vladimir Araujo and Abinew Ali Ayele and Pavan Baswani and Meriem Beloucif and Chris Biemann and Sofia Bourhim and Christine De Kock and Genet Shanko Dekebo and Oumaima Hourrane and Gopichand Kanumolu and Lokesh Madasu and Samuel Rutunda and Manish Shrivastava and Thamar Solorio and Nirmal Surange and Hailegnaw Getaneh Tilaye and Krishnapriya Vishnubhotla and Genta Winata and Seid Muhie Yimam and Saif M. Mohammad},
      year={2024},
      eprint={2402.08638},
      archivePrefix={arXiv},
      primaryClass={cs.CL},
      url={https://arxiv.org/abs/2402.08638}, 
}

@inproceedings{Bandarkar_2024,
   title={The Belebele Benchmark: a Parallel Reading Comprehension Dataset in 122 Language Variants},
   url={http://dx.doi.org/10.18653/v1/2024.acl-long.44},
   DOI={10.18653/v1/2024.acl-long.44},
   booktitle={Proceedings of the 62nd Annual Meeting of the Association for Computational Linguistics (Volume 1: Long Papers)},
   publisher={Association for Computational Linguistics},
   author={Bandarkar, Lucas and Liang, Davis and Muller, Benjamin and Artetxe, Mikel and Shukla, Satya Narayan and Husa, Donald and Goyal, Naman and Krishnan, Abhinandan and Zettlemoyer, Luke and Khabsa, Madian},
   year={2024},
   pages={749–775} }

@misc{thapliyal2022crossmodal3600massivelymultilingualmultimodal,
      title={Crossmodal-3600: A Massively Multilingual Multimodal Evaluation Dataset}, 
      author={Ashish V. Thapliyal and Jordi Pont-Tuset and Xi Chen and Radu Soricut},
      year={2022},
      eprint={2205.12522},
      archivePrefix={arXiv},
      primaryClass={cs.CV},
      url={https://arxiv.org/abs/2205.12522}, 
}

@inproceedings{Artetxe_2020,
   title={On the Cross-lingual Transferability of Monolingual Representations},
   url={http://dx.doi.org/10.18653/v1/2020.acl-main.421},
   DOI={10.18653/v1/2020.acl-main.421},
   booktitle={Proceedings of the 58th Annual Meeting of the Association for Computational Linguistics},
   publisher={Association for Computational Linguistics},
   author={Artetxe, Mikel and Ruder, Sebastian and Yogatama, Dani},
   year={2020},
   pages={4623–4637} }

@article{10.1162/tacl_a_00595,
    author = {Zhang, Xinyu and Thakur, Nandan and Ogundepo, Odunayo and Kamalloo, Ehsan and Alfonso-Hermelo, David and Li, Xiaoguang and Liu, Qun and Rezagholizadeh, Mehdi and Lin, Jimmy},
    title = {MIRACL: A Multilingual Retrieval Dataset Covering 18 Diverse Languages},
    journal = {Transactions of the Association for Computational Linguistics},
    volume = {11},
    pages = {1114-1131},
    year = {2023},
    month = {09},
    issn = {2307-387X},
    doi = {10.1162/tacl_a_00595},
    url = {https://doi.org/10.1162/tacl_a_00595},
    eprint = {https://direct.mit.edu/tacl/article-pdf/doi/10.1162/tacl_a_00595/2157340/tacl_a_00595.pdf},
}

@misc{williams2018broadcoveragechallengecorpussentence,
      title={A Broad-Coverage Challenge Corpus for Sentence Understanding through Inference}, 
      author={Adina Williams and Nikita Nangia and Samuel R. Bowman},
      year={2018},
      eprint={1704.05426},
      archivePrefix={arXiv},
      primaryClass={cs.CL},
      url={https://arxiv.org/abs/1704.05426}, 
}

@misc{bowman2015largeannotatedcorpuslearning,
      title={A large annotated corpus for learning natural language inference}, 
      author={Samuel R. Bowman and Gabor Angeli and Christopher Potts and Christopher D. Manning},
      year={2015},
      eprint={1508.05326},
      archivePrefix={arXiv},
      primaryClass={cs.CL},
      url={https://arxiv.org/abs/1508.05326}, 
}

@misc{nllbteam2022languageleftbehindscaling,
      title={No Language Left Behind: Scaling Human-Centered Machine Translation}, 
      author={NLLB-Team and Marta R. Costa-jussà and James Cross and Onur Çelebi and Maha Elbayad and Kenneth Heafield and Kevin Heffernan and Elahe Kalbassi and Janice Lam and Daniel Licht and Jean Maillard and Anna Sun and Skyler Wang and Guillaume Wenzek and Al Youngblood and Bapi Akula and Loic Barrault and Gabriel Mejia Gonzalez and Prangthip Hansanti and John Hoffman and Semarley Jarrett and Kaushik Ram Sadagopan and Dirk Rowe and Shannon Spruit and Chau Tran and Pierre Andrews and Necip Fazil Ayan and Shruti Bhosale and Sergey Edunov and Angela Fan and Cynthia Gao and Vedanuj Goswami and Francisco Guzmán and Philipp Koehn and Alexandre Mourachko and Christophe Ropers and Safiyyah Saleem and Holger Schwenk and Jeff Wang},
      year={2022},
      eprint={2207.04672},
      archivePrefix={arXiv},
      primaryClass={cs.CL},
      url={https://arxiv.org/abs/2207.04672}, 
}

@misc{li2025ssacometllmsoutperformlearned,
      title={SSA-COMET: Do LLMs Outperform Learned Metrics in Evaluating MT for Under-Resourced African Languages?}, 
      author={Senyu Li and Jiayi Wang and Felermino D. M. A. Ali and Colin Cherry and Daniel Deutsch and Eleftheria Briakou and Rui Sousa-Silva and Henrique Lopes Cardoso and Pontus Stenetorp and David Ifeoluwa Adelani},
      year={2025},
      eprint={2506.04557},
      archivePrefix={arXiv},
      primaryClass={cs.CL},
      url={https://arxiv.org/abs/2506.04557}, 
}

@misc{chen2024bgem3embeddingmultilingualmultifunctionality,
      title={BGE M3-Embedding: Multi-Lingual, Multi-Functionality, Multi-Granularity Text Embeddings Through Self-Knowledge Distillation}, 
      author={Jianlv Chen and Shitao Xiao and Peitian Zhang and Kun Luo and Defu Lian and Zheng Liu},
      year={2024},
      eprint={2402.03216},
      archivePrefix={arXiv},
      primaryClass={cs.CL},
      url={https://arxiv.org/abs/2402.03216}, 
}

@misc{zhang2025qwen3embeddingadvancingtext,
      title={Qwen3 Embedding: Advancing Text Embedding and Reranking Through Foundation Models}, 
      author={Yanzhao Zhang and Mingxin Li and Dingkun Long and Xin Zhang and Huan Lin and Baosong Yang and Pengjun Xie and An Yang and Dayiheng Liu and Junyang Lin and Fei Huang and Jingren Zhou},
      year={2025},
      eprint={2506.05176},
      archivePrefix={arXiv},
      primaryClass={cs.CL},
      url={https://arxiv.org/abs/2506.05176}, 
}

@misc{lee2025geminiembeddinggeneralizableembeddings,
      title={Gemini Embedding: Generalizable Embeddings from Gemini}, 
      author={Jinhyuk Lee and Feiyang Chen and Sahil Dua and Daniel Cer and Madhuri Shanbhogue and Iftekhar Naim and Gustavo Hernández Ábrego and Zhe Li and Kaifeng Chen and Henrique Schechter Vera and Xiaoqi Ren and Shanfeng Zhang and Daniel Salz and Michael Boratko and Jay Han and Blair Chen and Shuo Huang and Vikram Rao and Paul Suganthan and Feng Han and Andreas Doumanoglou and Nithi Gupta and Fedor Moiseev and Cathy Yip and Aashi Jain and Simon Baumgartner and Shahrokh Shahi and Frank Palma Gomez and Sandeep Mariserla and Min Choi and Parashar Shah and Sonam Goenka and Ke Chen and Ye Xia and Koert Chen and Sai Meher Karthik Duddu and Yichang Chen and Trevor Walker and Wenlei Zhou and Rakesh Ghiya and Zach Gleicher and Karan Gill and Zhe Dong and Mojtaba Seyedhosseini and Yunhsuan Sung and Raphael Hoffmann and Tom Duerig},
      year={2025},
      eprint={2503.07891},
      archivePrefix={arXiv},
      primaryClass={cs.CL},
      url={https://arxiv.org/abs/2503.07891}, 
}

@misc{adelani2023masakhanewsnewstopicclassification,
      title={MasakhaNEWS: News Topic Classification for African languages}, 
      author={David Ifeoluwa Adelani and Marek Masiak and Israel Abebe Azime and Jesujoba Alabi and Atnafu Lambebo Tonja and et. al},
      year={2023},
      eprint={2304.09972},
      archivePrefix={arXiv},
      primaryClass={cs.CL},
      url={https://arxiv.org/abs/2304.09972}, 
}

@misc{muhammad2023afrisentitwittersentimentanalysis,
      title={AfriSenti: A Twitter Sentiment Analysis Benchmark for African Languages}, 
      author={Shamsuddeen Hassan Muhammad and Idris Abdulmumin and Abinew Ali Ayele and Nedjma Ousidhoum and David Ifeoluwa Adelani and Seid Muhie Yimam and Ibrahim Sa'id Ahmad and Meriem Beloucif and Saif M. Mohammad and Sebastian Ruder and Oumaima Hourrane and Pavel Brazdil and Felermino Dário Mário António Ali and Davis David and Salomey Osei and Bello Shehu Bello and Falalu Ibrahim and Tajuddeen Gwadabe and Samuel Rutunda and Tadesse Belay and Wendimu Baye Messelle and Hailu Beshada Balcha and Sisay Adugna Chala and Hagos Tesfahun Gebremichael and Bernard Opoku and Steven Arthur},
      year={2023},
      eprint={2302.08956},
      archivePrefix={arXiv},
      primaryClass={cs.CL},
      url={https://arxiv.org/abs/2302.08956}, 
}

@misc{muhammad2025afrihatemultilingualcollectionhate,
      title={AfriHate: A Multilingual Collection of Hate Speech and Abusive Language Datasets for African Languages}, 
      author={Shamsuddeen Hassan Muhammad and Idris Abdulmumin and Abinew Ali Ayele and David Ifeoluwa Adelani and Ibrahim Said Ahmad and Saminu Mohammad Aliyu and Nelson Odhiambo Onyango and Lilian D. A. Wanzare and Samuel Rutunda and Lukman Jibril Aliyu and Esubalew Alemneh and Oumaima Hourrane and Hagos Tesfahun Gebremichael and Elyas Abdi Ismail and Meriem Beloucif and Ebrahim Chekol Jibril and Andiswa Bukula and Rooweither Mabuya and Salomey Osei and Abigail Oppong and Tadesse Destaw Belay and Tadesse Kebede Guge and Tesfa Tegegne Asfaw and Chiamaka Ijeoma Chukwuneke and Paul Röttger and Seid Muhie Yimam and Nedjma Ousidhoum},
      year={2025},
      eprint={2501.08284},
      archivePrefix={arXiv},
      primaryClass={cs.CL},
      url={https://arxiv.org/abs/2501.08284}, 
}

@misc{yu2025injongomulticulturalintentdetection,
      title={INJONGO: A Multicultural Intent Detection and Slot-filling Dataset for 16 African Languages}, 
      author={Hao Yu and Jesujoba O. Alabi and Andiswa Bukula and Jian Yun Zhuang and En-Shiun Annie Lee and Tadesse Kebede Guge and Israel Abebe Azime and Happy Buzaaba and Blessing Kudzaishe Sibanda and Godson K. Kalipe and Jonathan Mukiibi and Salomon Kabongo Kabenamualu and Mmasibidi Setaka and Lolwethu Ndolela and Nkiruka Odu and Rooweither Mabuya and Shamsuddeen Hassan Muhammad and Salomey Osei and Sokhar Samb and Juliet W. Murage and Dietrich Klakow and David Ifeoluwa Adelani},
      year={2025},
      eprint={2502.09814},
      archivePrefix={arXiv},
      primaryClass={cs.CL},
      url={https://arxiv.org/abs/2502.09814}, 
}

@inproceedings{federmann-etal-2022-ntrex,
    title = "{NTREX}-128 {--} News Test References for {MT} Evaluation of 128 Languages",
    author = "Federmann, Christian  and
      Kocmi, Tom  and
      Xin, Ying",
    editor = "Ahuja, Kabir  and
      Anastasopoulos, Antonios  and
      Patra, Barun  and
      Neubig, Graham  and
      Choudhury, Monojit  and
      Dandapat, Sandipan  and
      Sitaram, Sunayana  and
      Chaudhary, Vishrav",
    booktitle = "Proceedings of the First Workshop on Scaling Up Multilingual Evaluation",
    month = nov,
    year = "2022",
    address = "Online",
    publisher = "Association for Computational Linguistics",
    url = "https://aclanthology.org/2022.sumeval-1.4/",
    doi = "10.18653/v1/2022.sumeval-1.4",
    pages = "21--24"
}

@misc{wang2024improvingtextembeddingslarge,
      title={Improving Text Embeddings with Large Language Models}, 
      author={Liang Wang and Nan Yang and Xiaolong Huang and Linjun Yang and Rangan Majumder and Furu Wei},
      year={2024},
      eprint={2401.00368},
      archivePrefix={arXiv},
      primaryClass={cs.CL},
      url={https://arxiv.org/abs/2401.00368}, 
}

@misc{ojo2025afrobenchgoodlargelanguage,
      title={AfroBench: How Good are Large Language Models on African Languages?}, 
      author={Jessica Ojo and Odunayo Ogundepo and Akintunde Oladipo and Kelechi Ogueji and Jimmy Lin and Pontus Stenetorp and David Ifeoluwa Adelani},
      year={2025},
      eprint={2311.07978},
      archivePrefix={arXiv},
      primaryClass={cs.CL},
      url={https://arxiv.org/abs/2311.07978}, 
}

@misc{niyongabo2020kinnewskirnewsbenchmarkingcrosslingual,
      title={KINNEWS and KIRNEWS: Benchmarking Cross-Lingual Text Classification for Kinyarwanda and Kirundi}, 
      author={Rubungo Andre Niyongabo and Hong Qu and Julia Kreutzer and Li Huang},
      year={2020},
      eprint={2010.12174},
      archivePrefix={arXiv},
      primaryClass={cs.CL},
      url={https://arxiv.org/abs/2010.12174}, 
}

@misc{muhammad2022naijasentinigeriantwittersentiment,
      title={NaijaSenti: A Nigerian Twitter Sentiment Corpus for Multilingual Sentiment Analysis}, 
      author={Shamsuddeen Hassan Muhammad and David Ifeoluwa Adelani and Sebastian Ruder and Ibrahim Said Ahmad and Idris Abdulmumin and Bello Shehu Bello and Monojit Choudhury and Chris Chinenye Emezue and Saheed Salahudeen Abdullahi and Anuoluwapo Aremu and Alipio Jeorge and Pavel Brazdil},
      year={2022},
      eprint={2201.08277},
      archivePrefix={arXiv},
      primaryClass={cs.CL},
      url={https://arxiv.org/abs/2201.08277}, 
}

@misc{ponwitayarat2025seabedsoutheastasiaembedding,
      title={SEA-BED: Southeast Asia Embedding Benchmark}, 
      author={Wuttikorn Ponwitayarat and Raymond Ng and Jann Railey Montalan and Thura Aung and Jian Gang Ngui and Yosephine Susanto and William Tjhi and Panuthep Tasawong and Erik Cambria and Ekapol Chuangsuwanich and Sarana Nutanong and Peerat Limkonchotiwat},
      year={2025},
      eprint={2508.12243},
      archivePrefix={arXiv},
      primaryClass={cs.CL},
      url={https://arxiv.org/abs/2508.12243}, 
}

@misc{poświata2024plmtebpolishmassivetext,
      title={PL-MTEB: Polish Massive Text Embedding Benchmark}, 
      author={Rafał Poświata and Sławomir Dadas and Michał Perełkiewicz},
      year={2024},
      eprint={2405.10138},
      archivePrefix={arXiv},
      primaryClass={cs.CL},
      url={https://arxiv.org/abs/2405.10138}, 
}

@misc{ciancone2024mtebfrenchresourcesfrenchsentence,
      title={MTEB-French: Resources for French Sentence Embedding Evaluation and Analysis}, 
      author={Mathieu Ciancone and Imene Kerboua and Marion Schaeffer and Wissam Siblini},
      year={2024},
      eprint={2405.20468},
      archivePrefix={arXiv},
      primaryClass={cs.CL},
      url={https://arxiv.org/abs/2405.20468}, 
}

@misc{xiao2024cpackpackedresourcesgeneral,
      title={C-Pack: Packed Resources For General Chinese Embeddings}, 
      author={Shitao Xiao and Zheng Liu and Peitian Zhang and Niklas Muennighoff and Defu Lian and Jian-Yun Nie},
      year={2024},
      eprint={2309.07597},
      archivePrefix={arXiv},
      primaryClass={cs.CL},
      url={https://arxiv.org/abs/2309.07597}, 
}

@misc{wehrli2024germantextembeddingclustering,
      title={German Text Embedding Clustering Benchmark}, 
      author={Silvan Wehrli and Bert Arnrich and Christopher Irrgang},
      year={2024},
      eprint={2401.02709},
      archivePrefix={arXiv},
      primaryClass={cs.CL},
      url={https://arxiv.org/abs/2401.02709}, 
}

@misc{jmteb,
    author = {Li, Shengzhe and Ohagi, Masaya and Ri, Ryokan},
    title = {{J}{M}{T}{E}{B}: {J}apanese {M}assive {T}ext {E}mbedding {B}enchmark},
    howpublished = {\url{https://huggingface.co/datasets/sbintuitions/JMTEB}},
    year = {2024},
}

@misc{snegirev2025russianfocusedembeddersexplorationrumteb,
      title={The Russian-focused embedders' exploration: ruMTEB benchmark and Russian embedding model design}, 
      author={Artem Snegirev and Maria Tikhonova and Anna Maksimova and Alena Fenogenova and Alexander Abramov},
      year={2025},
      eprint={2408.12503},
      archivePrefix={arXiv},
      primaryClass={cs.CL},
      url={https://arxiv.org/abs/2408.12503}, 
}

@misc{zinvandi2025famtebmassivetextembedding,
      title={FaMTEB: Massive Text Embedding Benchmark in Persian Language}, 
      author={Erfan Zinvandi and Morteza Alikhani and Mehran Sarmadi and Zahra Pourbahman and Sepehr Arvin and Reza Kazemi and Arash Amini},
      year={2025},
      eprint={2502.11571},
      archivePrefix={arXiv},
      primaryClass={cs.CL},
      url={https://arxiv.org/abs/2502.11571}, 
}

@misc{vera2025embeddinggemmapowerfullightweighttext,
      title={EmbeddingGemma: Powerful and Lightweight Text Representations}, 
      author={Henrique Schechter Vera and Sahil Dua and Biao Zhang and Daniel Salz and et. al},
      year={2025},
      eprint={2509.20354},
      archivePrefix={arXiv},
      primaryClass={cs.CL},
      url={https://arxiv.org/abs/2509.20354}, 
}

@misc{li2023generaltextembeddingsmultistage,
      title={Towards General Text Embeddings with Multi-stage Contrastive Learning}, 
      author={Zehan Li and Xin Zhang and Yanzhao Zhang and Dingkun Long and Pengjun Xie and Meishan Zhang},
      year={2023},
      eprint={2308.03281},
      archivePrefix={arXiv},
      primaryClass={cs.CL},
      url={https://arxiv.org/abs/2308.03281}, 
}

@misc{neelakantan2022textcodeembeddingscontrastive,
      title={Text and Code Embeddings by Contrastive Pre-Training}, 
      author={Arvind Neelakantan and Tao Xu and Raul Puri and Alec Radford and Jesse Michael Han and Jerry Tworek and Qiming Yuan and Nikolas Tezak and Jong Wook Kim and Chris Hallacy and Johannes Heidecke and Pranav Shyam and Boris Power and Tyna Eloundou Nekoul and Girish Sastry and Gretchen Krueger and David Schnurr and Felipe Petroski Such and Kenny Hsu and Madeleine Thompson and Tabarak Khan and Toki Sherbakov and Joanne Jang and Peter Welinder and Lilian Weng},
      year={2022},
      eprint={2201.10005},
      archivePrefix={arXiv},
      primaryClass={cs.CL},
      url={https://arxiv.org/abs/2201.10005}, 
}

@misc{SFRAIResearch2024,
  title={SFR-Embedding-Mistral:Enhance Text Retrieval with Transfer Learning},
  author={Rui Meng and Ye Liu and Shafiq Rayhan Joty and Caiming Xiong and Yingbo Zhou and Semih Yavuz},
  howpublished={Salesforce AI Research Blog},
  year={2024},
  url={https://www.salesforce.com/blog/sfr-embedding/}
}

@misc{LinqAIResearch2024,
  title={Linq-Embed-Mistral:Elevating Text Retrieval with Improved GPT Data Through Task-Specific Control and Quality Refinement},
  author={Junseong Kim and Seolhwa Lee and Jihoon Kwon and Sangmo Gu and Yejin Kim and Minkyung Cho and Jy-yong Sohn and Chanyeol Choi},
  howpublished={Linq AI Research Blog},
  year={2024},
  url={https://getlinq.com/blog/linq-embed-mistral/}
}

@misc{muennighoff2024generative,
      title={Generative Representational Instruction Tuning}, 
      author={Niklas Muennighoff and Hongjin Su and Liang Wang and Nan Yang and Furu Wei and Tao Yu and Amanpreet Singh and Douwe Kiela},
      year={2024},
      eprint={2402.09906},
      archivePrefix={arXiv},
      primaryClass={cs.CL}
}

@article{li2023towards,
  title={Towards general text embeddings with multi-stage contrastive learning},
  author={Li, Zehan and Zhang, Xin and Zhang, Yanzhao and Long, Dingkun and Xie, Pengjun and Zhang, Meishan},
  journal={arXiv preprint arXiv:2308.03281},
  year={2023}
}

@article{hu2025kalm,
  title={KaLM-Embedding: Superior Training Data Brings A Stronger Embedding Model},
  author={Hu, Xinshuo and Shan, Zifei and Zhao, Xinping and Sun, Zetian and Liu, Zhenyu and Li, Dongfang and Ye, Shaolin and Wei, Xinyuan and Chen, Qian and Hu, Baotian and others},
  journal={arXiv preprint arXiv:2501.01028},
  year={2025}
}

@inproceedings{warner-etal-2025-smarter,
    title = "Smarter, Better, Faster, Longer: A Modern Bidirectional Encoder for Fast, Memory Efficient, and Long Context Finetuning and Inference",
    author = {Warner, Benjamin  and
      Chaffin, Antoine  and
      Clavi{\'e}, Benjamin  and
      Weller, Orion  and
      Hallstr{\"o}m, Oskar  and
      Taghadouini, Said  and
      Gallagher, Alexis  and
      Biswas, Raja  and
      Ladhak, Faisal  and
      Aarsen, Tom  and
      Adams, Griffin Thomas  and
      Howard, Jeremy  and
      Poli, Iacopo},
    editor = "Che, Wanxiang  and
      Nabende, Joyce  and
      Shutova, Ekaterina  and
      Pilehvar, Mohammad Taher",
    booktitle = "Proceedings of the 63rd Annual Meeting of the Association for Computational Linguistics (Volume 1: Long Papers)",
    month = jul,
    year = "2025",
    address = "Vienna, Austria",
    publisher = "Association for Computational Linguistics",
    url = "https://aclanthology.org/2025.acl-long.127/",
    doi = "10.18653/v1/2025.acl-long.127",
    pages = "2526--2547",
    ISBN = "979-8-89176-251-0"
}

@misc{marone2025mmbertmodernmultilingualencoder,
      title={mmBERT: A Modern Multilingual Encoder with Annealed Language Learning}, 
      author={Marc Marone and Orion Weller and William Fleshman and Eugene Yang and Dawn Lawrie and Benjamin Van Durme},
      year={2025},
      eprint={2509.06888},
      archivePrefix={arXiv},
      primaryClass={cs.CL},
      url={https://arxiv.org/abs/2509.06888}, 
}

@misc{conneau2020unsupervisedcrosslingualrepresentationlearning,
      title={Unsupervised Cross-lingual Representation Learning at Scale}, 
      author={Alexis Conneau and Kartikay Khandelwal and Naman Goyal and Vishrav Chaudhary and Guillaume Wenzek and Francisco Guzmán and Edouard Grave and Myle Ott and Luke Zettlemoyer and Veselin Stoyanov},
      year={2020},
      eprint={1911.02116},
      archivePrefix={arXiv},
      primaryClass={cs.CL},
      url={https://arxiv.org/abs/1911.02116}, 
}

@misc{reimers2019sentencebertsentenceembeddingsusing,
      title={Sentence-BERT: Sentence Embeddings using Siamese BERT-Networks}, 
      author={Nils Reimers and Iryna Gurevych},
      year={2019},
      eprint={1908.10084},
      archivePrefix={arXiv},
      primaryClass={cs.CL},
      url={https://arxiv.org/abs/1908.10084}, 
}

@misc{muhammad2025brighterbridginggaphumanannotated,
      title={BRIGHTER: BRIdging the Gap in Human-Annotated Textual Emotion Recognition Datasets for 28 Languages}, 
      author={Shamsuddeen Hassan Muhammad and Nedjma Ousidhoum and Idris Abdulmumin and Jan Philip Wahle and Terry Ruas and Meriem Beloucif and Christine de Kock and Nirmal Surange and Daniela Teodorescu and Ibrahim Said Ahmad and David Ifeoluwa Adelani and Alham Fikri Aji and Felermino D. M. A. Ali and Ilseyar Alimova and Vladimir Araujo and Nikolay Babakov and Naomi Baes and Ana-Maria Bucur and Andiswa Bukula and Guanqun Cao and Rodrigo Tufino Cardenas and Rendi Chevi and Chiamaka Ijeoma Chukwuneke and Alexandra Ciobotaru and Daryna Dementieva and Murja Sani Gadanya and Robert Geislinger and Bela Gipp and Oumaima Hourrane and Oana Ignat and Falalu Ibrahim Lawan and Rooweither Mabuya and Rahmad Mahendra and Vukosi Marivate and Alexander Panchenko and Andrew Piper and Charles Henrique Porto Ferreira and Vitaly Protasov and Samuel Rutunda and Manish Shrivastava and Aura Cristina Udrea and Lilian Diana Awuor Wanzare and Sophie Wu and Florian Valentin Wunderlich and Hanif Muhammad Zhafran and Tianhui Zhang and Yi Zhou and Saif M. Mohammad},
      year={2025},
      eprint={2502.11926},
      archivePrefix={arXiv},
      primaryClass={cs.CL},
      url={https://arxiv.org/abs/2502.11926}, 
}

@misc{pham2025vnmtebvietnamesemassivetext,
      title={VN-MTEB: Vietnamese Massive Text Embedding Benchmark}, 
      author={Loc Pham and Tung Luu and Thu Vo and Minh Nguyen and Viet Hoang},
      year={2025},
      eprint={2507.21500},
      archivePrefix={arXiv},
      primaryClass={cs.CL},
      url={https://arxiv.org/abs/2507.21500}, 
}

@inproceedings{
enevoldsen2025mmteb,
title={{MMTEB}: Massive Multilingual Text Embedding Benchmark},
author={Kenneth Enevoldsen and Isaac Chung and Imene Kerboua and M{\'a}rton Kardos and Ashwin Mathur and David Stap and Jay Gala and Wissam Siblini and Dominik Krzemi{\'n}ski and Genta Indra Winata and Saba Sturua and Saiteja Utpala and Mathieu Ciancone and Marion Schaeffer and Diganta Misra and Shreeya Dhakal and Jonathan Rystr{\o}m and Roman Solomatin and {\"O}mer Veysel {\c{C}}a{\u{g}}atan and Akash Kundu and Martin Bernstorff and Shitao Xiao and Akshita Sukhlecha and Bhavish Pahwa and Rafa{\l} Po{\'s}wiata and Kranthi Kiran GV and Shawon Ashraf and Daniel Auras and Bj{\"o}rn Pl{\"u}ster and Jan Philipp Harries and Lo{\"\i}c Magne and Isabelle Mohr and Dawei Zhu and Hippolyte Gisserot-Boukhlef and Tom Aarsen and Jan Kostkan and Konrad Wojtasik and Taemin Lee and Marek Suppa and Crystina Zhang and Roberta Rocca and Mohammed Hamdy and Andrianos Michail and John Yang and Manuel Faysse and Aleksei Vatolin and Nandan Thakur and Manan Dey and Dipam Vasani and Pranjal A Chitale and Simone Tedeschi and Nguyen Tai and Artem Snegirev and Mariya Hendriksen and Michael G{\"u}nther and Mengzhou Xia and Weijia Shi and Xing Han L{\`u} and Jordan Clive and Gayatri K and Maksimova Anna and Silvan Wehrli and Maria Tikhonova and Henil Shalin Panchal and Aleksandr Abramov and Malte Ostendorff and Zheng Liu and Simon Clematide and Lester James Validad Miranda and Alena Fenogenova and Guangyu Song and Ruqiya Bin Safi and Wen-Ding Li and Alessia Borghini and Federico Cassano and Lasse Hansen and Sara Hooker and Chenghao Xiao and Vaibhav Adlakha and Orion Weller and Siva Reddy and Niklas Muennighoff},
booktitle={The Thirteenth International Conference on Learning Representations},
year={2025},
url={https://openreview.net/forum?id=zl3pfz4VCV}
}

@inproceedings{bandarkar-etal-2024-belebele,
    title = "The Belebele Benchmark: a Parallel Reading Comprehension Dataset in 122 Language Variants",
    author = "Bandarkar, Lucas  and
      Liang, Davis  and
      Muller, Benjamin  and
      Artetxe, Mikel  and
      Shukla, Satya Narayan  and
      Husa, Donald  and
      Goyal, Naman  and
      Krishnan, Abhinandan  and
      Zettlemoyer, Luke  and
      Khabsa, Madian",
    editor = "Ku, Lun-Wei  and
      Martins, Andre  and
      Srikumar, Vivek",
    booktitle = "Proceedings of the 62nd Annual Meeting of the Association for Computational Linguistics (Volume 1: Long Papers)",
    month = aug,
    year = "2024",
    address = "Bangkok, Thailand",
    publisher = "Association for Computational Linguistics",
    url = "https://aclanthology.org/2024.acl-long.44/",
    doi = "10.18653/v1/2024.acl-long.44",
    pages = "749--775"
}

@article{alabi2025charting,
  title={Charting the Landscape of African NLP: Mapping Progress and Shaping the Road Ahead},
  author={Alabi, Jesujoba O and Hedderich, Michael A and Adelani, David Ifeoluwa and Klakow, Dietrich},
  journal={arXiv preprint arXiv:2505.21315},
  year={2025}
}

@inproceedings{adelani-etal-2025-irokobench,
    title = "{I}roko{B}ench: A New Benchmark for {A}frican Languages in the Age of Large Language Models",
    author = "Adelani, David Ifeoluwa  and
      Ojo, Jessica  and
      Azime, Israel Abebe  and
      Zhuang, Jian Yun  and
      Alabi, Jesujoba Oluwadara  and
      He, Xuanli  and
      Ochieng, Millicent  and
      Hooker, Sara  and
      Bukula, Andiswa  and
      Lee, En-Shiun Annie  and
      Chukwuneke, Chiamaka Ijeoma  and
      Buzaaba, Happy  and
      Sibanda, Blessing Kudzaishe  and
      Kalipe, Godson Koffi  and
      Mukiibi, Jonathan  and
      Kabongo Kabenamualu, Salomon  and
      Yuehgoh, Foutse  and
      Setaka, Mmasibidi  and
      Ndolela, Lolwethu  and
      Odu, Nkiruka  and
      Mabuya, Rooweither  and
      Osei, Salomey  and
      Muhammad, Shamsuddeen Hassan  and
      Samb, Sokhar  and
      Guge, Tadesse Kebede  and
      Sherman, Tombekai Vangoni  and
      Stenetorp, Pontus",
    editor = "Chiruzzo, Luis  and
      Ritter, Alan  and
      Wang, Lu",
    booktitle = "Proceedings of the 2025 Conference of the Nations of the Americas Chapter of the Association for Computational Linguistics: Human Language Technologies (Volume 1: Long Papers)",
    month = apr,
    year = "2025",
    address = "Albuquerque, New Mexico",
    publisher = "Association for Computational Linguistics",
    url = "https://aclanthology.org/2025.naacl-long.139/",
    doi = "10.18653/v1/2025.naacl-long.139",
    pages = "2732--2757",
    ISBN = "979-8-89176-189-6"
}

@inproceedings{belay-etal-2025-evaluating,
    title = "Evaluating the Capabilities of Large Language Models for Multi-label Emotion Understanding",
    author = "Belay, Tadesse Destaw  and
      Azime, Israel Abebe  and
      Ayele, Abinew Ali  and
      Sidorov, Grigori  and
      Klakow, Dietrich  and
      Slusallek, Philip  and
      Kolesnikova, Olga  and
      Yimam, Seid Muhie",
    editor = "Rambow, Owen  and
      Wanner, Leo  and
      Apidianaki, Marianna  and
      Al-Khalifa, Hend  and
      Eugenio, Barbara Di  and
      Schockaert, Steven",
    booktitle = "Proceedings of the 31st International Conference on Computational Linguistics",
    month = jan,
    year = "2025",
    address = "Abu Dhabi, UAE",
    publisher = "Association for Computational Linguistics",
    url = "https://aclanthology.org/2025.coling-main.237/",
    pages = "3523--3540"
}

@inproceedings{fitzgerald-etal-2023-massive,
    title = "{MASSIVE}: A 1{M}-Example Multilingual Natural Language Understanding Dataset with 51 Typologically-Diverse Languages",
    author = "FitzGerald, Jack  and
      Hench, Christopher  and
      Peris, Charith  and
      Mackie, Scott  and
      Rottmann, Kay  and
      Sanchez, Ana  and
      Nash, Aaron  and
      Urbach, Liam  and
      Kakarala, Vishesh  and
      Singh, Richa  and
      Ranganath, Swetha  and
      Crist, Laurie  and
      Britan, Misha  and
      Leeuwis, Wouter  and
      Tur, Gokhan  and
      Natarajan, Prem",
    editor = "Rogers, Anna  and
      Boyd-Graber, Jordan  and
      Okazaki, Naoaki",
    booktitle = "Proceedings of the 61st Annual Meeting of the Association for Computational Linguistics (Volume 1: Long Papers)",
    month = jul,
    year = "2023",
    address = "Toronto, Canada",
    publisher = "Association for Computational Linguistics",
    url = "https://aclanthology.org/2023.acl-long.235/",
    doi = "10.18653/v1/2023.acl-long.235",
    pages = "4277--4302"
}

@inproceedings{rei-etal-2020-comet,
    title = "{COMET}: A Neural Framework for {MT} Evaluation",
    author = "Rei, Ricardo  and
      Stewart, Craig  and
      Farinha, Ana C  and
      Lavie, Alon",
    editor = "Webber, Bonnie  and
      Cohn, Trevor  and
      He, Yulan  and
      Liu, Yang",
    booktitle = "Proceedings of the 2020 Conference on Empirical Methods in Natural Language Processing (EMNLP)",
    month = nov,
    year = "2020",
    address = "Online",
    publisher = "Association for Computational Linguistics",
    url = "https://aclanthology.org/2020.emnlp-main.213/",
    doi = "10.18653/v1/2020.emnlp-main.213",
    pages = "2685--2702"
}

\appendix

\section{Appendix}
\label{sec:appendix}
\subsection{Task Categories}
\label{subsec:afrimteb-families}

AfriMTEB follows the MTEB taxonomy and groups tasks into eight families. Table~\ref{tab:afrimteb-families} lists the families and the datasets included in the \emph{Full} suite. 


\paragraph{Bitext Mining.}
Given two sentence sets from different languages, the task is to identify translation pairs. Embeddings are used to compute similarities and find the best match for each sentence.
\paragraph{Pair Classification.}
This task involves a pair of input sentences with a binary or categorical relationship label (e.g., entailment vs.\ contradiction). Predictions are based on embedding similarity.
\paragraph{Classification.}
Single-text classification where each input is mapped to one label among several categories (e.g., topic, sentiment, hate speech, or language ID). A linear classifier is trained on top of embeddings.
\paragraph{Multi-label Classification.}
Texts may be assigned multiple labels simultaneously (e.g., emotions). A multi-label classifier is applied on top of embeddings to handle overlapping categories.
\paragraph{Clustering.}
Given a collection of texts, the task is to group them into clusters that correspond to gold categories. Embeddings are clustered with algorithms such as $k$-means.
\paragraph{Semantic Text Similarity.}
This task measures the degree of semantic similarity between sentence pairs, either within or across languages, based on embedding similarity.
\paragraph{Retrieval.}
Given a query, the task is to retrieve relevant documents from a large corpus. Both queries and documents are embedded, and similarity scores determine ranking.
\paragraph{Reranking.}
Given a query and a candidate set of documents, the goal is to rank the candidates by relevance using embeddings. This task focuses on fine-grained ranking quality.

\subsection{Dataset Descriptions}
\label{subsec:afrimteb-dataset-descriptions}

\subsubsection{Bitext Mining datasets}

\paragraph{Flores \citep{goyal-etal-2022-flores}.}
FLORES is a widely used multilingual parallel corpus designed to support evaluation of cross-lingual transfer. It consists of sentence-aligned translations across a large number of languages, including many African languages. In AfriMTEB, FLORES is cast as a bitext mining task: given a sentence in one language, the model must retrieve its correct translation from a pool of candidate sentences in another language using embedding similarity. This task evaluates the ability of embeddings to align semantically equivalent content across languages.

\paragraph{NTREX \citep{federmann-etal-2022-ntrex}.}
NTREX is a multilingual parallel dataset derived from professionally translated content, covering a broad set of languages. Similar to FLORES, it is framed as a bitext mining task in AfriMTEB, where embeddings are used to retrieve aligned translations. NTREX complements FLORES by providing different domains and translation characteristics, allowing evaluation of robustness across translation styles.

\paragraph{BibleNLP \citep{kann-2024-massively}.}
BibleNLP consists of verse-aligned Bible translations across many languages, including low-resource African languages. Although the domain is religious text, the strict verse alignment makes it well suited for evaluating cross-lingual semantic alignment. In AfriMTEB, the task is to retrieve the correct verse translation given a source verse, testing whether embeddings capture meaning consistently across languages despite domain specificity.

\paragraph{NollySenti \citep{shode2023nollysentileveragingtransferlearning}.}
NollySenti originates as a sentiment analysis dataset for Nigerian languages but is additionally repurposed in AfriMTEB as a bitext-style mining task by leveraging aligned or parallel content. This dataset evaluates whether embeddings trained primarily for semantic similarity can also recover aligned text across languages in a more informal, social-media-driven domain.

\paragraph{Tatoeba \citep{tiedemann2020tatoebatranslationchallenge}.}
The Tatoeba dataset contains sentence-aligned translations created by a community of contributors and covers a wide range of languages. In AfriMTEB, Tatoeba is used for bitext mining, where models must identify translation pairs based on embedding similarity. The dataset is noisier than professionally curated corpora.

\subsubsection{Pair Classification datasets}

\paragraph{XNLI \citep{conneau2018xnlievaluatingcrosslingualsentence}.}
XNLI is a cross-lingual natural language inference benchmark derived from MultiNLI. Each example consists of a premise–hypothesis sentence pair labeled as entailment, contradiction, or neutral. In AfriMTEB, XNLI is treated as a pair classification task, where embeddings of the two sentences are combined to predict the relation label, evaluating whether embeddings preserve fine-grained semantic relations.

\paragraph{AfriXNLI \citep{adelani-etal-2025-irokobench}.}
AfriXNLI extends the XNLI framework to additional African languages. It follows the same premise–hypothesis structure and label space as XNLI, but focuses on languages that are largely absent from existing NLI benchmarks. This dataset allows AfriMTEB to evaluate whether multilingual embeddings generalize NLI-style reasoning to African languages.

\subsubsection{Classification datasets}

\paragraph{SIB200Classification \citep{adelani2024sib200simpleinclusivebig}.}
SIB200Classification is derived from the SIB-200 benchmark and formulates topic classification with a reduced set of coarse-grained categories. Each document is assigned exactly one topic label. In AfriMTEB, it is used to evaluate general topic classification across a large number of languages with balanced label distributions.

\paragraph{SIB200\_14Classes \citep{adelani2024sib200simpleinclusivebig}.}
SIB200\_14Classes is a more fine-grained variant of topic classification from SIB-200, where texts are assigned to one of fourteen topic categories. This dataset is more challenging due to the larger label space and semantic overlap between topics. It is included in AfriMTEB to assess how well embeddings separate closely related topical categories across languages.

\paragraph{MasakhaNEWS \citep{adelani2023masakhanewsnewstopicclassification}.}
MasakhaNEWS is a multilingual African news classification benchmark covering several African languages. Each news article is labeled with a topic such as politics, sports, or business. In AfriMTEB, it serves as a core news topic classification dataset and reflects realistic, domain-specific text encountered in African media.

\paragraph{TswanaNews, SiswatiNews \citep{madodonga2023izindabatindzabamachinelearningnews}, SwahiliNews, IsiZuluNews \citep{madodonga2023izindabatindzabamachinelearningnews}, KinNews \citep{niyongabo2020kinnewskirnewsbenchmarkingcrosslingual}.}
These datasets are language-specific news classification benchmarks for Setswana, SiSwati, Swahili, isiZulu, and Kinyarwanda respectively. Each dataset follows the same formulation as MasakhaNEWS, with articles labeled by topic. KinNews is newly incorporated in AfriMTEB to extend news classification coverage to Kinyarwanda, ensuring broader geographic representation.

\paragraph{NaijaSenti \citep{muhammad2022naijasentinigeriantwittersentiment}.}
NaijaSenti is a sentiment analysis dataset for Nigerian languages, primarily focused on informal and social-media-style text. Each example is labeled with sentiment polarity (e.g., positive, negative, neutral). In AfriMTEB, it evaluates whether embeddings capture affective meaning in informal language varieties.

\paragraph{AfriSenti \citep{muhammad2023afrisentitwittersentimentanalysis}.}
AfriSenti is a multilingual African sentiment dataset covering multiple languages and domains. Compared to NaijaSenti, it includes a broader linguistic scope and more diverse text sources. It is used to assess cross-lingual sentiment classification performance.

\paragraph{MultilingualSentiment \citep{muennighoff2023mtebmassivetextembedding}.}
This dataset is a general multilingual sentiment benchmark included to provide additional coverage beyond African-specific datasets. It allows comparison between African-language performance and more widely studied multilingual sentiment settings.

\paragraph{AfriHate \citep{muhammad2025afrihatemultilingualcollectionhate}.}
AfriHate is a hate speech and offensive language classification dataset designed specifically for African languages. Each text is labeled according to whether it contains hate or abusive content. Its inclusion addresses a gap in prior benchmarks, where African languages were largely absent from hate speech evaluation.

\paragraph{LanguageClassification, SouthAfricanLangClassification \citep{south-african-language-identification}, AfriSentiLangClassification \citep{muhammad2023afrisentitwittersentimentanalysis}.}
These datasets are used for language identification (LID). The task is to predict the language of a given text. SouthAfricanLangClassification focuses on closely related South African languages, while AfriSentiLangClassification is derived from sentiment data and tests LID under informal text conditions.

\paragraph{MassiveIntent \citep{ousidhoum2024semrel2024collectionsemantictextual}.}
MassiveIntent comes from the MASSIVE dataset and consists of short user utterances labeled with intent categories (e.g., request, command). It evaluates whether embeddings encode intent-level semantics across languages.

\paragraph{InjongoIntent \citep{yu2025injongomulticulturalintentdetection}.}
InjongoIntent is an intent classification dataset targeting African languages. It mirrors the structure of MassiveIntent but focuses on underrepresented languages and locally relevant intents, providing a more realistic evaluation for African conversational systems.

\paragraph{MassiveScenario \citep{ousidhoum2024semrel2024collectionsemantictextual}.}
MassiveScenario is another subset of the MASSIVE dataset, where utterances are labeled by scenario or domain (e.g., travel, finance). In AfriMTEB, it is treated as a standard single-label classification task.

\subsubsection{Multi-label Classification datasets}

\paragraph{EmotionAnalysisPlus \citep{muhammad2025semeval2025task11bridging}.}
EmotionAnalysisPlus is a multi-label emotion classification dataset where each text may express multiple emotions simultaneously (e.g., joy, anger, sadness). Unlike sentiment analysis, this task requires modeling overlapping affective states. It is included in AfriMTEB to evaluate multi-label prediction capabilities in African languages.

\subsubsection{Semantic Text Similarity datasets}

\paragraph{SemRel24STS \citep{ousidhoum2024semrel2024collectionsemantictextual}.}
SemRel24STS is a semantic textual similarity dataset where sentence pairs are annotated with graded similarity scores. Models are evaluated by correlating embedding-based similarity with human judgments. This dataset tests fine-grained semantic sensitivity rather than categorical decisions.

\subsubsection{Retrieval datasets}

\paragraph{Belebele \citep{Bandarkar_2024}.}
Belebele is a multilingual retrieval benchmark derived from reading comprehension data. Given a query (often a question), the task is to retrieve the most relevant passage from a set of candidates. It evaluates cross-lingual retrieval performance under realistic QA-style conditions.

\paragraph{MIRACL and MIRACLRetrievalHardNegatives \citep{10.1162/tacl_a_00595}.}
MIRACL is a large-scale multilingual information retrieval benchmark covering many languages. Queries are paired with large document collections, and models must retrieve relevant passages. The Hard Negatives variant augments the task with challenging non-relevant passages, making ranking more difficult and discriminative.

\paragraph{MrTidy \citep{zhang2021mrtydimultilingualbenchmark}, XQuAD \citep{Artetxe_2020}, XM3600T2I \citep{thapliyal2022crossmodal3600massivelymultilingualmultimodal}.}
These datasets extend retrieval evaluation to different domains and modalities. MrTidy and XQuAD focus on text-based retrieval, while XM3600T2I evaluates cross-lingual text-to-image retrieval. Together, they broaden the retrieval evaluation beyond standard document search.

\subsubsection{Clustering datasets}

\paragraph{SIB200ClusteringFast \citep{adelani2024sib200simpleinclusivebig}.}
This dataset is derived from SIB-200 and evaluates topic clustering. Texts must be grouped into clusters corresponding to gold topic labels, using only embedding similarity and clustering algorithms such as $k$-means.

\paragraph{MasakhaNEWSClusteringP2P and MasakhaNEWSClusteringS2S~\citep{adelani2023masakhanewsnewstopicclassification}.}
These datasets formulate clustering tasks from MasakhaNEWS articles using different clustering protocols. They evaluate whether embeddings induce meaningful topical structure in African news text.

\subsubsection{Reranking datasets}

\paragraph{MIRACLReranking \citep{10.1162/tacl_a_00595}.}
MIRACLReranking is a fine-grained ranking task built on MIRACL. Given a query and a small set of candidate passages, the goal is to rerank them by relevance. This task focuses on precise ordering rather than coarse retrieval, testing the discriminative power of embeddings.

\begin{table}[h]
\centering
\footnotesize
\begin{tabular}{ll}
\toprule
\textbf{Parameter} & \textbf{Value} \\
\midrule
Training epochs & 1 \\
Batch size (per device) & 8 \\
Group size & 8 \\
Learning rate & 1e-5 \\
Warmup ratio & 0.1 \\
Max query length & 512 \\
Max passage length & 512 \\
Padding multiple & 8 \\
Knowledge distillation & True (KL divergence) \\
Negatives cross-device & Enabled \\
Embedding normalization & True (L2) \\
Sentence pooling & Mean pooling \\
Temperature & 0.02 \\
Precision & FP16 \\
\bottomrule
\end{tabular}
\caption{Key training configurations used in fine-tuning.}
\label{tab:train-config}
\end{table}

\subsection{Baseline Model Descriptions}
\label{subsec:baseline}

\noindent
\textbf{mmBERT-base}~\citep{marone2025mmbertmodernmultilingualencoder} is a modern multilingual encoder trained on more than $3$T tokens covering $>$1{,}800 languages, extending the ModernBERT~\citep{warner-etal-2025-smarter} (fast encoder with long context) to the multilingual regime. The training recipe introduces curriculum-style annealed language learning, inverse masking, and temperature-based sampling to emphasize low-resource languages while retaining strong performance on high-resource ones. Reported results show that mmBERT-base surpasses prior multilingual encoders such as XLM-R on standard NLU and retrieval benchmarks, approaching the English-only ModernBERT on GLUE despite being trained predominantly on non-English data. Inference follows standard encoder usage (mean pooling over last hidden states and L2 normalization). 

\vspace{2mm}
\noindent
\textbf{KaLM-Embedding}~\citep{hu2025kalm} is a multilingual embedding family that prioritizes training-data quality over sheer scale, combining (i) persona-based synthetic examples distilled from LLMs, (ii) ranking-consistency filtering, and (iii) semi-homogeneous task batching for efficient contrastive learning. Many public checkpoints are built on compact Qwen2 backbones (e.g., $\sim$0.5B) and instruction-tuned variants for downstream retrieval and semantic similarity. Technical reports describe v1.5/v2 updates with improved data curation and training strategy, yielding strong MTEB performance for their size.

\vspace{2mm}
\noindent
\textbf{Qwen3-Embedding (0.6B / 4B / 8B)}~\citep{zhang2025qwen3embeddingadvancingtext} is a purpose-built series of dense encoders for text embeddings and reranking, offered in 0.6B/4B/8B sizes with a 32k token context window and coverage of 100+\ languages. The models are instruction-aware and support flexible output dimensionalities (via MRL-style prefix truncation), with typical maximum embedding sizes of 1{,}024 (0.6B), 2{,}560 (4B), and 4{,}096 (8B); matching reranker models are available at each size. Training leverages Qwen3 LLMs both as backbones and as data synthesizers across domains and languages, improving robustness for retrieval and reranking workloads. Inference uses mean pooling and L2 normalization; common deployment stacks expose a user-selectable output dimensionality for storage/latency trade-offs.

\vspace{2mm}
\noindent
\textbf{BGE-M3}~\citep{chen2024bgem3embeddingmultilingualmultifunctionality} is a versatile embedding model unifying three capabilities in a single encoder: Multi-Functionality (dense, multi-vector, and sparse retrieval), Multi-Linguality (100+\ languages), and Multi-Granularity (robust from short queries to long documents, up to $\sim$8{,}192 tokens). The training pipeline uses self-knowledge distillation across retrieval functions to align representations and enables hybrid retrieval without switching models. It is widely used as a strong multilingual baseline for retrieval, clustering, and classification.

\vspace{2mm}
\noindent
\textbf{mE5-Large} and \textbf{mE5-Large-Instruct}~\citep{wang2024multilinguale5textembeddings} are multilingual members of the E5 family built on XLM-RoBERTa-large \citep{conneau2020unsupervisedcrosslingualrepresentationlearning}, trained with a two-stage recipe that first performs weakly supervised contrastive pre-training on roughly one billion multilingual text pairs and then supervised fine-tuning on curated embedding tasks; the instruction variant further formats supervision with concise task instructions to specialize representations for retrieval and related tasks. Both use a 24-layer encoder that produces 1,024-dimensional vectors and inherit broad (~100 language) coverage from the XLM-RoBERTa backbone. At inference time they follow the E5 prompt conventions (e.g., query / passage style inputs), apply mean pooling over the last hidden states, and L2-normalize the output, yielding strong performance on retrieval, semantic similarity, clustering, and classification benchmarks in both monolingual and cross-lingual settings. 

\vspace{2mm}
\noindent
\textbf{gte-Qwen2-7B-instruct}~\citep{li2023towards} is a 7B-parameter instruction-tuned General Text Embedding model built on Qwen2-7B, targeting high-quality multilingual embeddings with a 32k context window. At release, it reported leading scores on MTEB English/Chinese subsets, reflecting a training mixture that combines instruction-style contrastive objectives and curated negatives. It is used as a strong large-model baseline for retrieval, reranking, and semantic similarity.

\vspace{2mm}
\noindent
\textbf{GritLM-7B}~\citep{muennighoff2024generative} unifies generation and embeddings in one model via Generative Representational Instruction Tuning (GRIT) . A custom modeling component adds bidirectional attention paths so that the same backbone can function as a strong encoder for embeddings without sacrificing generative performance \citep{muennighoff2024generative}. The paper reports SOTA-level MTEB results for 7B-class open models alongside strong generative benchmarks, demonstrating that a single model can excel at both modalities \citep{muennighoff2024generative}.

\vspace{2mm}
\noindent
\textbf{Linq-Embed-Mistral}~\citep{LinqAIResearch2024} is a Mistral-7B–based embedding model developed with task-tailored data crafting, filtering, and hard-negative mining to improve retrieval quality. The technical report and model card document leading MTEB retrieval scores at release (e.g., average $\sim$68.2 on 56 datasets; retrieval score $\sim$60.2), achieved through homogeneous task ordering and mixed-task fine-tuning strategies. It is frequently used as a competitive 7B encoder baseline for dense retrieval.

\vspace{2mm}
\noindent
\textbf{SFR-Embedding-Mistral}~\citep{SFRAIResearch2024} applies transfer learning on top of E5-mistral-7b-instruct and Mistral-7B-v0.1, with additional multi-task training and optimized negative sampling aimed at retrieval tasks. Public materials position it as a top-performing 7B embedding model for search, clustering, and classification workloads.

\vspace{2mm}
\noindent
\textbf{E5 Mistral 7B Instruct}~\citep{wang2024improvingtextembeddingslarge} initializes the E5 instruction-tuning recipe from Mistral-7B-v0.1, producing a 7B encoder specialized for instruction-aware embeddings . As with other E5 variants, inference benefits from task instructions plus ``\texttt{query:}/\texttt{passage:}'' prefixes, and mean pooling with L2 normalization is used for final vectors. It serves both as a competitive baseline and as a foundation for further transfer learning (e.g., SFR-Embedding-Mistral).

\vspace{2mm}
\noindent
\textbf{Gemini Embedding-001}~\citep{lee2025geminiembeddinggeneralizableembeddings} is Google’s multilingual text-embedding model available via the Gemini API and Vertex AI, trained with Matryoshka Representation Learning (MRL) so that leading vector prefixes remain useful at smaller dimensions . The default output is 3{,}072 dimensions, but APIs allow setting output dimensionality (e.g., 1{,}536 or 768) with minimal quality loss, enabling flexible storage/latency trade-offs. The model supports 100+\ languages and has been a strong performer on multilingual MTEB since its early releases.

\subsection{Detailed Statistics of Machine Translation Quality}
\label{sec:ssa_comet_samples}

To assess the quality of the machine translation data used in AfriMTEB, we rely on automatic evaluation with \textbf{SSA-COMET} \citep{li2025ssacometllmsoutperformlearned}. SSA-COMET is a recently released metric trained on \textsc{SSA-MTE}, a large-scale human-annotated evaluation dataset covering 13 African language pairs with over 63,000 sentence-level judgments. Compared to earlier African-focused metrics such as AfriCOMET, SSA-COMET provides stronger correlation with human ratings and better robustness in low-resource settings. 

\begin{table}[h]
\centering
\footnotesize
\setlength{\tabcolsep}{6pt}
\begin{tabular}{lrrr}
\toprule
\textbf{Language} & \textbf{0.67} & \textbf{0.75} & \textbf{0.80} \\
\midrule
Amharic (amh\_Ethi) & 44,166 & 4,617 & 281 \\
Oromo (gaz\_Latn)   & 73,846 & 14,598 & 2,818 \\
Hausa (hau\_Latn)   & 31,799 & 5,851 & 1,056 \\
Igbo (ibo\_Latn)    & 16,778 & 1,279 & 136 \\
Kinyarwanda (kin\_Latn) & 81,003 & 4,867 & 337 \\
Swahili (swh\_Latn) & 116,338 & 21,996 & 1,849 \\
Xhosa (xho\_Latn)   & 18,143 & 2,007 & 351 \\
Yoruba (yor\_Latn)  & 31,377 & 3,316 & 411 \\
Zulu (zul\_Latn)    & 20,229 & 1,535 & 224 \\
\midrule
\textbf{Total} & \textbf{433,629} & \textbf{60,066} & \textbf{7,463} \\
\bottomrule
\end{tabular}
\caption{Number of translation pairs retained after filtering with SSA-COMET at three thresholds. Lower thresholds yield more data, while stricter thresholds retain fewer but higher-quality examples.}
\label{tab:ssa-comet-thresholds}
\end{table}

\begin{table}[h]
\centering
\footnotesize
\setlength{\tabcolsep}{6pt}
\begin{tabular}{lrrr}
\toprule
\textbf{Language} & \textbf{0.67} & \textbf{0.75} & \textbf{0.80} \\
\midrule
Amharic (amh\_Ethi) & 44,166 & 4,617 & 281 \\
Oromo (gaz\_Latn)   & 73,846 & 14,598 & 2,818 \\
Hausa (hau\_Latn)   & 31,799 & 5,851 & 1,056 \\
Igbo (ibo\_Latn)    & 16,778 & 1,279 & 136 \\
Kinyarwanda (kin\_Latn) & 81,003 & 4,867 & 337 \\
Swahili (swh\_Latn) & 116,338 & 21,996 & 1,849 \\
Xhosa (xho\_Latn)   & 18,143 & 2,007 & 351 \\
Yoruba (yor\_Latn)  & 31,377 & 3,316 & 411 \\
Zulu (zul\_Latn)    & 20,229 & 1,535 & 224 \\
\midrule
Kongo (kon\_Latn)   & -- & 18,030 & -- \\
Kabuverdiano (kea\_Latn) & -- & 810 & -- \\
Somali (som\_Latn)  & -- & 4,203 & -- \\
Twi (twi\_Latn)     & -- & 6,090 & -- \\
Sotho (sot\_Latn)   & -- & 20,190 & -- \\
Tsonga (tso\_Latn)  & -- & 43,371 & -- \\
Swati (ssw\_Latn)   & -- & 4,062 & -- \\
Lingala (lin\_Latn) & -- & 20,091 & -- \\
Shona (sna\_Latn)   & -- & 23,025 & -- \\
Northern Sotho (nso\_Latn) & -- & 33,912 & -- \\
Plateau Malagasy (plt\_Latn) & -- & 11,736 & -- \\
Tswana (tsn\_Latn)  & -- & 35,481 & -- \\
Afrikaans (afr\_Latn) & -- & 19,872 & -- \\
Nyanja (nya\_Latn)  & -- & 53,367 & -- \\
Egyptian Arabic (arz\_Arab) & -- & 6,960 & -- \\
\midrule
\textbf{Total} & \textbf{433,629} & \textbf{405631} & \textbf{7,463} \\
\bottomrule
\end{tabular}
\caption{Updated counts of translation pairs retained after filtering with SSA-COMET at three thresholds.}
\label{tab:ssa-comet-thresholds-updated}
\end{table}

From Table~\ref{tab:ssa-comet-thresholds}, we observe a clear trade-off between dataset size and quality. At a relaxed threshold of 0.67, over 430k sentence pairs are retained, ensuring broad coverage across all nine target languages. Increasing the threshold to 0.75 reduces the pool to around 60k examples, striking a balance between filtering noise and preserving sufficient training data. At the strictest cutoff of 0.80, only 7.5k pairs remain, indicating that high-quality translations are relatively scarce. Language-level differences are also evident: Swahili, Oromo, and Kinyarwanda consistently contribute the largest number of pairs, while Igbo and Amharic see the sharpest reductions under stricter filtering, reflecting variation in MT quality across languages.

\subsection{Detailed AfriMTEB-Lite Results}
\label{sec:detailed_afrimteblite}
\autoref{tab:afrimteb_task_wise_results} shows the comprehensive results across all evaluated datasets and languages in \afrimteblite{}.

\subsection{Statistical Significance}
\label{app:statistical_significance}

To quantify the robustness of the observed improvements of AfriE5 over mE5, we conduct a paired bootstrap significance analysis on both AfriMTEB-Lite and AfriMTEB-Full.

For AfriMTEB-Lite, we treat each language within a task as one observation and compute paired differences
$\Delta = \text{AfriE5} - \text{mE5}$ per task--language cell. We then perform 10{,}000 paired bootstrap resamples over languages and report (i) the mean difference $\Delta$ (in absolute points), (ii) 95\% confidence intervals obtained via the percentile method, and (iii) one-sided $p$-values for the null hypothesis $H_0 : \Delta \le 0$.
The detailed results are shown in Table~\ref{tab:lite_significance}.

\begin{table}[t]
\centering
\caption{Statistical Significance on AfriMTEB-Lite Tasks. 
Mean performance difference $\Delta = \text{AfriE5} - \text{mE5}$ (in \%), 95\% confidence intervals (CI), and one-sided $p$-values ($H_0: \Delta \le 0$) estimated via paired bootstrap over languages.}
\label{tab:lite_significance}
\resizebox{\columnwidth}{!}{%
\begin{tabular}{lcccc}
\toprule
Task & $\Delta$ & 95\% CI & $p$-value & $n_\text{langs}$ \\
\midrule
AfriHate Classification   & +0.17 & $[-1.17, +1.38]$ & 0.390 & 9 \\
AfriSenti Classification  & +3.74 & $[+2.08, +5.40]$ & $\boldsymbol{< 0.001}$ & 7 \\
News Classification       & +0.64 & $[-0.06, +1.30]$ & $\boldsymbol{0.035}$ & 8 \\
AfriXNLI                  & +4.50 & $[+3.82, +5.16]$ & $\boldsymbol{< 0.001}$ & 9 \\
Emotion Analysis          & +1.28 & $[+0.30, +2.25]$ & $\boldsymbol{0.004}$ & 9 \\
Flores Bitext Mining      & -0.17 & $[-0.26, -0.06]$ & 0.998 & 9 \\
Injongo Intent            & -0.04 & $[-1.41, +1.45]$ & 0.540 & 9 \\
NTREX Bitext Mining       & +0.53 & $[-0.14, +1.54]$ & 0.109 & 9 \\
SIB-200 (14 Classes)      & +4.16 & $[+2.58, +5.75]$ & $\boldsymbol{< 0.001}$ & 9 \\
SIB-200 Classification    & +0.84 & $[+0.17, +1.59]$ & $\boldsymbol{0.004}$ & 9 \\
SIB-200 Clustering        & +1.76 & $[+0.77, +2.91]$ & $\boldsymbol{< 0.001}$ & 9 \\
Belebele Retrieval        & +2.00 & $[+1.59, +2.41]$ & $\boldsymbol{< 0.001}$ & 9 \\
\midrule
Overall (Macro)           & +1.62 & $[+0.79, +2.57]$ & $\boldsymbol{< 0.001}$ & 12 \\
Overall (Micro)           & +1.59 & $[+1.17, +2.02]$ & $\boldsymbol{< 0.001}$ & 105 \\
\bottomrule
\end{tabular}%
}
\end{table}

On AfriMTEB-Lite, 9 out of 12 tasks show statistically significant improvements at $p < 0.05$, and both macro and micro averages are significant. The largest gains appear on AfriXNLI, SIB-200 (14 classes), SIB-200 clustering, AfriSenti, and Belebele retrieval, while Flores bitext mining shows a small regression.

For AfriMTEB-Full, we group task--language cells into the eight categories reported in the main paper (bitext mining, classification, clustering, multilabel classification, pair classification, reranking, retrieval, and STS). Within each category we again compute paired differences per cell and apply the same 10{,}000-resample paired bootstrap procedure. Table~\ref{tab:full_category_significance} reports the mean difference, 95\% confidence intervals, one-sided $p$-values, and the number of cells per category.

\begin{table}[t]
\centering
\caption{Statistical Significance by Category (AfriMTEB-Full). 
Mean performance difference $\Delta = \text{AfriE5} - \text{mE5}$ averaged across all tasks/languages within each category. $p$-values are computed via paired bootstrap.}
\label{tab:full_category_significance}
\resizebox{\columnwidth}{!}{%
\begin{tabular}{lcccc}
\toprule
Category & $\Delta$ & 95\% CI & $p$-value & $n_\text{cells}$ \\
\midrule
Bitext Mining             & -0.17 & $[-0.36, +0.04]$ & 0.949 & 95 \\
Classification            & -0.16 & $[-1.09, +0.67]$ & 0.622 & 194 \\
Clustering                & +1.02 & $[-1.87, +3.76]$ & 0.235 & 26 \\
Multilabel Classification & +1.22 & $[+0.56, +1.88]$ & $\boldsymbol{< 0.001}$ & 14 \\
Pair Classification       & +4.19 & $[+3.47, +4.78]$ & $\boldsymbol{< 0.001}$ & 17 \\
Reranking                 & +2.13 & $[+1.58, +2.66]$ & $\boldsymbol{< 0.001}$ & 2 \\
Retrieval                 & +1.37 & $[+0.95, +1.75]$ & $\boldsymbol{< 0.001}$ & 36 \\
STS                       & -1.04 & $[-3.02, +0.98]$ & 0.865 & 7 \\
\midrule
Overall (Macro)           & +1.07 & $[+0.08, +2.15]$ & $\boldsymbol{0.015}$ & 8 \\
Overall (Micro)           & +0.29 & $[-0.22, +0.76]$ & 0.122 & 391 \\
\bottomrule
\end{tabular}%
}
\end{table}

AfriE5 significantly outperforms mE5 in four out of eight categories (multilabel classification, pair classification, reranking, retrieval), while bitext mining and STS show small, non-significant regressions. The overall macro average is significantly positive, whereas the micro average is not, largely because the large classification category shows only small, non-significant gains.

\subsection{Multiple Seeds Training}
\label{app:multiple_seeds}

To assess the stability of AfriE5 under our single-GPU training recipe, we additionally train three instances of AfriE5 with different random seeds (42, 123, 456), keeping all other hyperparameters and data fixed. Table~\ref{tab:afromteb_dataset_results} reports AfriMTEB-Lite results for mE5, the primary AfriE5 run reported in the main paper, and the three additional seed runs.

\begin{table*}[ht]
  \centering
  \footnotesize
  \setlength{\tabcolsep}{4pt}
  \resizebox{\textwidth}{!}{
  \begin{tabular}{@{}l*{13}{r}@{}}
    \toprule
    & \multicolumn{3}{c}{}%
    & \multicolumn{1}{c}{} &
    & \multicolumn{2}{c}{\textbf{Btxt}} &
    & \multicolumn{1}{c}{\textbf{}}%
    & \multicolumn{3}{c}{\textbf{SIB-200}}%
    & \multicolumn{1}{c}{}\\
    \cmidrule(lr){7-8} \cmidrule(lr){11-13}
    \textbf{Model} &
    \textbf{Hate} & \textbf{Senti} & \textbf{NLI} &
    \textbf{Retrvl} & \textbf{Emo} &
    \textbf{Flores} & \textbf{NTREX} &
    \textbf{Intent} & \textbf{News} &
    \textbf{14Classes} & \textbf{Class} & \textbf{Clust} &
    \textbf{Avg.} \\
    \midrule

    \m{mE5-large-instruct} &
    51.5 & 47.0 & 64.5 &
    75.7 & 31.5 &
    \textbf{91.4} & 91.5 &
    75.5 & 78.8 &
    22.0 & 71.2 & 43.9 &
    62.0 \\

    \m{AfriE5-large-instruct} &
    \textbf{51.7} & 50.7 & 69.0 &
    77.7 & \textbf{32.8} &
    91.2 & \textbf{92.0} &
    75.4 & 79.5 &
    \textbf{26.2} & 72.0 & 45.7 &
    \textbf{63.7} \\

    \midrule

    \m{seed42} &
    51.5 & 50.4 & 68.5 &
    \textbf{77.8} & 32.7 &
    91.3 & 91.9 &
    75.1 & \textbf{79.7} &
    26.1 & \textbf{72.1} & \textbf{45.9} &
    63.6 \\

    \m{seed123} &
    51.4 & \textbf{51.0} & \textbf{69.0} &
    77.6 & 32.7 &
    90.8 & 91.5 &
    75.4 & 79.5 &
    26.0 & 71.9 & 45.7 &
    63.5 \\

    \m{seed456} &
    51.5 & 50.4 & 68.6 &
    77.7 & 32.7 &
    \textbf{91.4} & 92.0 &
    \textbf{75.6} & 79.5 &
    25.5 & 71.9 & 45.6 &
    63.5 \\

    \bottomrule
  \end{tabular}}
  \vspace{-2mm}
  \caption{\textbf{AfriMTEB-Lite results.} Average performance of embedding models across nine African languages (\texttt{AMH, GAZ, HAU, IBO, KIN, SWA, XHO, YOR, ZUL}) on 12 tasks. Columns report task-level averages, and the final column gives the unweighted macro average across tasks. Best scores per column are highlighted in bold.}
  \label{tab:afromteb_dataset_results}
\end{table*}

Across the three additional seeds, the overall AfriMTEB-Lite macro average varies only between 63.5 and 63.6 (vs.\ 63.7 for the main AfriE5 run). Per-task scores are also highly consistent, with differences typically well below one absolute point and no systematic reversals of the performance pattern relative to mE5. This confirms that our adaptation procedure is stable with respect to random initialization and data ordering, even under a modest compute budget (single A100, one epoch over the filtered training set).

\begin{table*}[t]
\footnotesize
\centering
\scalebox{0.9}{
\begin{tabular}{lrrrrrrrrr|r}
\textbf{Dataset} & \textbf{amh} & \textbf{gaz} & \textbf{hau} & \textbf{ibo} & \textbf{kin} & \textbf{swa} & \textbf{xho} & \textbf{yor} & \textbf{zul} & \textbf{Avg.} \\
\toprule
\textbf{AfriHateClassification} &  &  &  &  &  &  &  &  &  &  \\
bge-m3 & 52.52 & 52.19 & 47.32 & 56.03 & 51.71 & 64.81 & 36.85 & 50.77 & 38.49 & 50.08 \\
\rowcolor{Gray}
gemini embedding 001 & 54.31 & 52.31 & 52.4 & 63.47 & 57.75 & 74.07 & 38.94 & 56.3 & 45.3 & 54.98 \\
mE5-large-instruct & 50.78 & 51.92 & 42.95 & 57.16 & 54.85 & 67.12 & 40.18 & 51.56 & 47.03 & 51.51 \\
AfriE5-large-instruct & 52.76 & 53.79 & 44.77 & 58.95 & 56.47 & 64.94 & 37.8 & 51.67 & 43.93 & 51.68 \\
\midrule
\textbf{AfriSentiClassification} &  &  &  &  &  &  &  &  &  &  \\
bge-m3 & 48.45 & 34.8 & 68.83 & 54.43 & 46.12 & 44.52 & -- & 38.21 & -- & 47.91 \\
\rowcolor{Gray}
gemini embedding 001 & 59.75 & 34.35 & 75.02 & 61.65 & 58.58 & 45.94 & -- & 41.25 & -- & 53.79 \\
mE5-large-instruct & 44.07 & 35.8 & 68.64 & 49.52 & 50.19 & 41.95 & -- & 38.91 & -- & 47.01 \\
AfriE5-large-instruct & 51.41 & 35.7 & 71.78 & 53.67 & 53.98 & 44.13 & -- & 44.57 & -- & 50.75 \\
\midrule
\textbf{NewsClassification} &  &  &  &  &  &  &  &  &  &  \\
bge-m3 & 84.73 & 74.62 & 78.23 & 64.23 & 49.4 & 73.8 & 77.85 & 79.05 & -- & 72.74 \\
gemini embedding 001 & 84.41 & 82.09 & 77.14 & 68.9 & 58.58 & 71.95 & 87.31 & 84.21 & -- & 76.82 \\
mE5-large-instruct & 87.82 & 79.54 & 80.71 & 77.59 & 57.27 & 79.43 & 85.29 & 83.07 & -- & 78.84 \\
\rowcolor{Gray}
AfriE5-large-instruct & 89.52 & 81.75 & 81.52 & 78.03 & 58.31 & 78.15 & 85.59 & 82.99 & -- & 79.48 \\
\midrule
\textbf{AfriXNLI} &  &  &  &  &  &  &  &  &  &  \\
bge-m3 & 75.64 & 66.39 & 69.67 & 64.33 & 65.27 & 73.88 & 69.41 & 64.03 & 69.87 & 68.72 \\
\rowcolor{Gray}
gemini embedding 001 & 81.32 & 66.8 & 78.57 & 78.05 & 73.79 & 78.34 & 75.42 & 71.72 & 73.38 & 75.27 \\
mE5-large-instruct & 66.85 & 62.96 & 63.51 & 65.05 & 62.01 & 68.09 & 66.66 & 61.89 & 63.7 & 64.52 \\
AfriE5-large-instruct & 72.28 & 68.57 & 67.89 & 70.36 & 64.68 & 72.1 & 72.41 & 65.43 & 67.52 & 69.03 \\
\midrule
\textbf{EmotionAnalysisPlus} &  &  &  &  &  &  &  &  &  &  \\
bge-m3 & 20.99 & 28.54 & 24.92 & 23.49 & 33.9 & 37.14 & 21.52 & 40.58 & 33.86 & 29.44 \\
gemini embedding 001 & 28.38 & 30.8 & 36.03 & 30.71 & 39.52 & 39.67 & 27.76 & 45.66 & 40.6 & 35.46 \\
mE5-large-instruct & 22.27 & 30.41 & 27.03 & 27.59 & 34.27 & 39.89 & 24.17 & 40.88 & 37.05 & 31.51 \\
\rowcolor{Gray}
AfriE5-large-instruct & 24.07 & 30.74 & 28.37 & 30.18 & 36.21 & 38.8 & 28.13 & 42.2 & 36.35 & 32.78 \\
\midrule
\textbf{FloresBitextMining} &  &  &  &  &  &  &  &  &  &  \\
bge-m3 & 83.73 & 71.4 & 82.71 & 72.78 & 76.63 & 86.13 & 81.99 & 65.33 & 82.24 & 78.1 \\
gemini embedding 001 & 91.3 & 77.8 & 90.01 & 87.9 & 90.13 & 91.43 & 90.37 & 83.99 & 90.28 & 88.13 \\
\rowcolor{Gray}
mE5-large-instruct & 92.58 & 88.51 & 91.5 & 91.65 & 92.05 & 93.22 & 92.58 & 87.88 & 92.72 & 91.41 \\
\rowcolor{Gray}
AfriE5-large-instruct & 92.37 & 88.67 & 91.32 & 91.32 & 91.83 & 92.91 & 92.29 & 87.92 & 92.51 & 91.24 \\
\midrule
\textbf{InjongoIntent} &  &  &  &  &  &  &  &  &  &  \\
bge-m3 & 85.84 & 61.19 & 86.5 & 71.86 & 65.66 & 90.36 & 74.62 & 74.11 & 68.05 & 75.35 \\
\rowcolor{Gray}
gemini embedding 001 & 89.5 & 65.23 & 95.25 & 83.14 & 77.72 & 93.64 & 87.38 & 84.28 & 79.22 & 83.93 \\
mE5-large-instruct & 80.86 & 64.89 & 85.05 & 74.45 & 65.95 & 80.17 & 78.11 & 79.38 & 70.25 & 75.46 \\
AfriE5-large-instruct & 82.84 & 62.31 & 85.95 & 72.3 & 65.2 & 84.42 & 79.67 & 77.73 & 68.36 & 75.42 \\
\midrule
\textbf{NTREXBitextMining} &  &  &  &  &  &  &  &  &  &  \\
bge-m3 & 79.36 & 60.9 & 86.3 & 81.64 & 79.04 & 93.91 & 83.76 & 72.93 & 86.07 & 80.43 \\
gemini embedding 001 & 87.2 & 65.69 & 88.16 & 87.03 & 81.02 & 94.49 & 86.33 & 77.45 & 90.35 & 84.19 \\
mE5-large-instruct & 92.07 & 73.8 & 94.81 & 94.57 & 92.65 & 97.45 & 93.82 & 89.26 & 95.15 & 91.51 \\
\rowcolor{Gray}
AfriE5-large-instruct & 92.87 & 78.04 & 94.67 & 94.63 & 91.83 & 97.43 & 93.88 & 89.67 & 95.37 & 92.04 \\
\midrule
\textbf{SIB200-14Classes} &  &  &  &  &  &  &  &  &  &  \\
bge-m3 & 16.05 & 6.08 & 13.59 & 6.85 & 9.23 & 16.92 & 9.83 & 4.08 & 9.31 & 10.22 \\
gemini embedding 001 & 18.27 & 8.13 & 18.58 & 16.24 & 23.23 & 20.03 & 18.46 & 13.64 & 23.07 & 17.74 \\
mE5-large-instruct & 21.24 & 12.61 & 22.11 & 21.49 & 23.8 & 32.44 & 21.63 & 17.23 & 25.51 & 22.01 \\
\rowcolor{Gray}
AfriE5-large-instruct & 30.21 & 13.28 & 27.48 & 25.91 & 29.02 & 33.05 & 27.06 & 20.71 & 28.82 & 26.17 \\
\midrule
\textbf{SIB200Classification} &  &  &  &  &  &  &  &  &  &  \\
bge-m3 & 64.9 & 44.22 & 60.85 & 51.16 & 53.44 & 67.99 & 56.41 & 46.73 & 55.07 & 55.64 \\
gemini embedding 001 & 71.72 & 57.15 & 69.65 & 69.84 & 73.69 & 73.4 & 71.01 & 65 & 71.3 & 69.2 \\
mE5-large-instruct & 72.87 & 57.09 & 71.27 & 73.7 & 74.68 & 76.63 & 74.1 & 66.05 & 74.14 & 71.17 \\
\rowcolor{Gray}
AfriE5-large-instruct & 75.47 & 59.82 & 72.57 & 73.69 & 75.16 & 76.46 & 73.8 & 67.1 & 74.04 & 72.01 \\
\midrule
\textbf{SIB200ClusteringFast} &  &  &  &  &  &  &  &  &  &  \\
bge-m3 & 29.82 & 10.84 & 22.91 & 15.01 & 18.84 & 34.49 & 21.51 & 13.4 & 22.02 & 20.98 \\
gemini embedding 001 & 39.18 & 20.92 & 40.94 & 37 & 37.29 & 38.3 & 34.57 & 27.63 & 35.68 & 34.61 \\
mE5-large-instruct & 48.96 & 30.47 & 44.47 & 46.11 & 47.42 & 50.32 & 46.65 & 36.03 & 45.04 & 43.94 \\
\rowcolor{Gray}
AfriE5-large-instruct & 54.35 & 31.61 & 44.87 & 46.56 & 48.65 & 49.95 & 49.53 & 38.31 & 47.52 & 45.71 \\
\midrule
 \textbf{BelebeleRetrieval}         &       &       &       &       &       &       &       &       &       &             \\
 BGE-m3                    & 79.99 & 58.88 & 75.48 & 61.26 & 65.58 & 87.92 & 69.13 & 60.84 & 69.59 & 69.85 \\
 \rowcolor{Gray}
 Gemini Embedding          & 91.07 & 68.23 & 86.46 & 80.26 & 86.01 & 92.65 & 86.33 & 75    & 86.61 & 83.62 \\
 mE5-Large-Instruct        & 81.08 & 66.34 & 77.54 & 73.72 & 76.93 & 86.97 & 77.63 & 62.97 & 77.97 & 75.68 \\
 AfriE5-Large-Instruct     & 83.34 & 69.42 & 78.71 & 75.71 & 78.34 & 87.99 & 80.16 & 65.25 & 80.26 & 77.69 \\
\bottomrule
\end{tabular}
}
\caption{\textbf{Task-wise and per-language Performance.} Comparison of the selected models across all African languages for each task.}
\label{tab:afrimteb_task_wise_results}
\end{table*}

\end{document}